\def\arxivversion{}
\PassOptionsToPackage{preprint}{tmlr}
\documentclass[10pt]{article}
\usepackage{tmlr}
\usepackage[utf8]{inputenc}
\usepackage[T1]{fontenc}
\usepackage{amsmath,amssymb,amsfonts}
\usepackage{graphicx,booktabs,multirow,longtable}
\usepackage{algorithm,algpseudocode}
\usepackage{placeins,listings,xcolor}
\usepackage{hyperref,url}
\providecommand{\selectlanguage}[1]{}
\hypersetup{colorlinks=true,linkcolor=blue,citecolor=blue,urlcolor=blue,
 pdftitle={Scientific discovery as meta-optimization: a combinatorial optimization case study},
 pdfauthor={Anonymous},pdfsubject={TMLR submission}}
\lstdefinestyle{prompt}{basicstyle=\footnotesize\ttfamily,frame=single,
 columns=fullflexible,keepspaces=true,breaklines=true,breakatwhitespace=false,
 aboveskip=0.8em,belowskip=0.5em}
\title{Scientific discovery as meta-optimization: a combinatorial optimization case study}
\author{\name Anonymous Authors}
\ifdefined\arxivversion
\hypersetup{
 pdfauthor={Yuan-Hang Zhang, Chesson Sipling, Massimiliano Di Ventra},
 pdfsubject={arXiv preprint}}
\author{
\name Yuan-Hang Zhang \email yuz092@ucsd.edu \\
\addr Department of Physics, University of California San Diego, La Jolla, CA 92093
\AND
\name Chesson Sipling \email csipling@ucsd.edu \\
\addr Department of Physics, University of California San Diego, La Jolla, CA 92093
\AND
\name Massimiliano Di Ventra \email diventra@physics.ucsd.edu \\
\addr Department of Physics, University of California San Diego, La Jolla, CA 92093
}

\fi
\begin{document}
\maketitle
\begin{abstract}
Scientific discovery is fundamentally an optimization problem, defined by a vast ``state space'' of theories and experiments, and an evaluation criterion based on quality, novelty, and validity. Large language models (LLMs) have enabled automated exploration of this state space, but we argue that simultaneous modification of the evaluation criteria is equally important. Here, we propose formalizing research as {\it meta-optimization}, where the optimization objective itself is also being optimized. Our key contribution is ``consensus objective aggregation,'' where a portfolio of LLM-generated objective functions is combined via correlation-weighted voting, allowing the evaluation criterion to evolve alongside the solutions. We apply this framework to algorithm discovery for planted 3-SAT problems based on digital memcomputing machines. The leading solver from the consensus search reduces median integration steps by more than $38\times$ relative to a tuned baseline at 2,560 variables. Objective ablations show that a well-chosen fixed expert objective can also discover strong solvers, while consensus supplies a stable ranking that mitigates reward hacking behaviors. By bringing objective engineering into the discovery loop, this framework offers a practical route toward more adaptable automated research.
\end{abstract}

\section{Introduction}
%%%%%%%%%%%%%%%%%%%%%%%%%%%%%%%%%%%%%%%%%%%%%%%%%%%%%%%%%%%%%%%%%%%%%%%%%%%%%%%

At its core, scientific research is an optimization process over a vast space of theories, experiments, and implementations~\citep{OUPbook}. Every published result corresponds to a local optimum in this ``research space,'' found through a slow, noisy sampling process carried out by humans. Evaluating a single idea may take months, and mistakes are common. Recently, large language models (LLMs) have started to automate this research process at scales far exceeding individual researchers, covering the entire loop from generating hypotheses, writing code, designing experiments, to analyzing results and writing academic papers~\citep{lu_ai_2024,yamada_ai_2025,gottweis_ai_2025,boiko_autonomous_2023,mitchener_kosmos_2025}.

The early results are striking. The AI Scientist~\citep{lu_ai_2024} and its successor~\citep{yamada_ai_2025} produce complete research papers across machine learning subfields. Google's AI co-scientist~\citep{gottweis_ai_2025} generates and debates biomedical hypotheses through a tournament-style loop. FunSearch~\citep{romera-paredes_mathematical_2024}, AlphaEvolve~\citep{novikov_alphaevolve_2025}, and SATLUTION~\citep{yu_autonomous_2025} pair LLMs with evolutionary search to push the state of the art on algorithm discovery. AI systems now tackle materials discovery~\citep{merchant_scaling_2023}, protein design~\citep{wang_swarms_2025}, chemical synthesis~\citep{boiko_autonomous_2023}, and the recovery of physical laws~\citep{han_physagent_2025}. Tree-search methods, including Monte Carlo Tree Search (MCTS), have been coupled with LLMs for heuristic design~\citep{zheng_monte_2025,wang_planning_2025,wang_automated_2025} and mathematical reasoning~\citep{zhang_accessing_2024,yao_tree_2023}. For surveys of this rapidly expanding area, see Refs.~\citep{wei_ai_2025,zheng_automation_2025,luo_llm4sr_2025,eger_transforming_2025}.

A more fundamental question, though, runs beneath all of these efforts: What should these systems actually optimize? The objectives guiding automated discovery, ranging from hand-crafted benchmarks, proxy scores, to LLM-generated evaluation functions, are imperfect measures for genuine research progress. This is Goodhart's law in action, well known from reinforcement learning: a proxy can lose its value as a measure when it becomes the target of optimization~\citep{goodhart_problems_1984}. Optimizing against proxy objectives invites reward hacking, in which solutions score well on the metric while failing at the underlying goal~\citep{skalse_defining_2022,su_end_2026}. Objective specification has been flagged as a central difficulty for AI-driven discovery~\citep{du_accelerating_2025}, and recent work on evolving objectives~\citep{du_accelerating_2025,lu_discovery_2025} has begun to address it---though usually by swapping one objective for the next in sequence.

In real scientific research, the goals are often unclear or evolving. A project may start with something like ``develop an efficient algorithm'' or ``discover a potent drug,'' but as understanding builds, the criteria for success shift. Goals are typically plural---accuracy, explanatory power, robustness, cost, novelty, societal impact---and may resist any single stable ordering. It is possible to define a meta-objective and optimize the objective function itself, but this simply pushes the same specification and gaming problem up one level. Therefore, we treat objectives as evolving, uncertain, and multi-dimensional, working with portfolios rather than individual metrics.

In this work, we then propose \emph{consensus objective aggregation} as a robust mechanism for {\it meta-optimization} in automated scientific discovery. Instead of maintaining a single evolving objective, the framework generates many objective functions, each encoding a different aspect of solution quality, and aggregates them through correlation-weighted voting. We compute pairwise Kendall's~$\tau$ rank correlations \citep{kendall_new_1938} across all objectives, weight each one by its median agreement with the rest (clamped to be non-negative) times an exponential age decay factor, and produce a consensus ranking via a weighted Borda count \citep{emerson_original_2013}.

Since each proxy objective is an imperfect attempt to capture the underlying research goal, agreement provides a useful signal for combining them. Correlation weighting reduces the influence of objectives that disagree with the portfolio, while age decay allows the ranking to adapt as objectives are added. A meta-agent can further adjust their influence using the research context. Together, these components turn objective engineering into part of the discovery loop.

This consensus mechanism sits inside a multi-agent system that iterates in a closed loop, where objective evolution is steered by a meta-agent and carried out by an objective agent, while actual research is outlined by the planner agent and implemented by the designer agent. We apply the framework to algorithm discovery for random planted 3-SAT problems~\citep{barthel_hiding_2002} using digital memcomputing machines (DMMs)~\citep{diventra_memcomputing_2022,traversa_polynomialtime_2017,bearden_efficient_2020,sipling_phasespace_2026}. The combined search archive contains 414 solver designs and 42 objectives. On fresh instances with 2,560 variables, a consensus-discovered solver reduces median integration steps by more than $38\times$ relative to a separately tuned baseline under the same memory cap. Additional search runs show how objective choice affects progress, with consensus offering a practical way to refine evaluation criteria during discovery.

%%%%%%%%%%%%%%%%%%%%%%%%%%%%%%%%%%%%%%%%%%%%%%%%%%%%%%%%%%%%%%%%%%%%%%%%%%%%%%%
\section{Results}
%%%%%%%%%%%%%%%%%%%%%%%%%%%%%%%%%%%%%%%%%%%%%%%%%%%%%%%%%%%%%%%%%%%%%%%%%%%%%%%

%---------------------------------------
\subsection{Framework Overview}
%---------------------------------------

\begin{figure}[!t]
\centering
\includegraphics[width=0.9\linewidth]{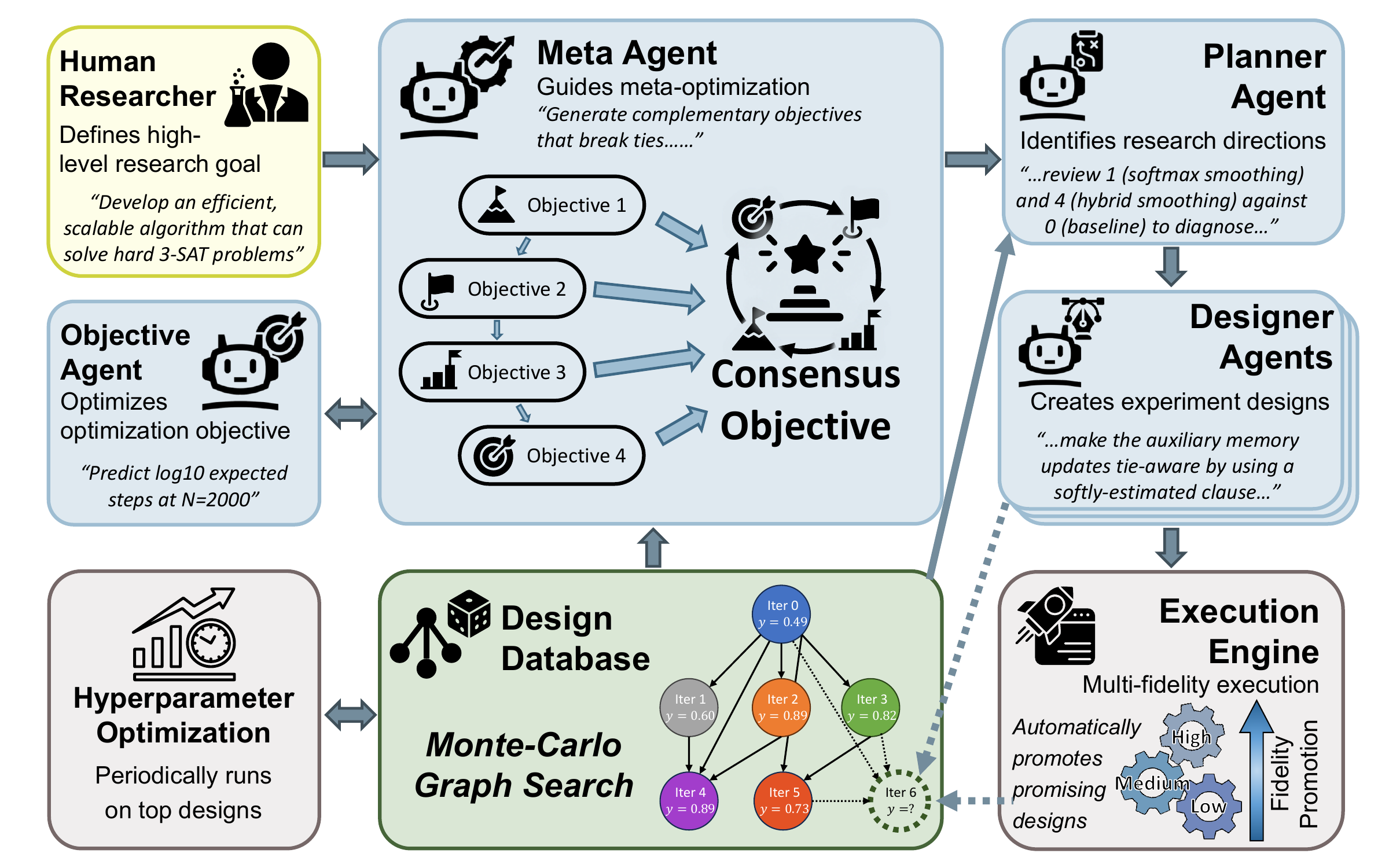}
\caption{\textbf{Framework overview.} The system consists of four LLM agents in an iterative cycle. Starting from a high-level human-designed research goal, the \textit{meta-agent} sets the research strategy, guiding objective generation and analyzing objective quality. The \textit{objective agent} proposes proxy objective functions reflecting different aspects of solution quality; these feed into a \textit{consensus objective} that aggregates rankings through correlation-weighted voting. The \textit{planner agent} uses Monte Carlo Graph Search (MCGS) under the consensus objective to identify strategic research directions. The \textit{designer agent} turns those directions into concrete implementations, which are tested by a \textit{multi-fidelity execution engine} that allocates computational budget to the most promising designs. A \textit{hyperparameter optimization} module periodically tunes the leading design's parameters.}
\label{fig:architecture}
\end{figure}

Our framework has four LLM agents in an iterative cycle (Fig.~\ref{fig:architecture}). Starting from a high-level goal defined by the human researcher, each iteration proceeds through the following stages:

\textbf{Meta-agent.} The meta-agent receives the human-defined research goal and steers the overall research strategy. Periodically, it analyzes existing objectives via Kendall's~$\tau$ correlations (Sec.~\ref{sec:consensus}), providing strategic guidance to the objective agent and assigning weight multipliers that can amplify useful objectives or suppress harmful ones.

\textbf{Objective agent.} Acting on the meta-agent's assessment, the objective agent generates new proxy objective functions mapping experiment results to scalar quality scores. Each objective is meant to capture a different aspect of solution quality. All objectives are scored on all designs and aggregated into a single consensus ranking through Kendall's~$\tau$ correlation-weighted voting with age decay (Sec.~\ref{sec:consensus} and Sec.~\ref{sec:consensus_methods}).

\textbf{Planner agent.} The planner receives the MCGS-ranked design list (scored by the consensus objective) along with the full experiment history. It looks for patterns among successes and failures and outputs several strategic research directions, with references for each direction to build upon.

\textbf{Designer agent.} For each direction the planner proposes, the designer writes new solver code and runs it through a multi-fidelity experiment schedule~\citep{peherstorfer_survey_2018}. Designs advance to higher fidelity levels by rule-based criteria relative to the current population, so that computational budget concentrates on the strongest candidates. The heteroscedastic evolutionary Bayesian optimization (HEBO) algorithm~\citep{cowen-rivers_hebo_2022} periodically tunes hyperparameters on the best untuned design.

This iterative cycle allows solutions and evaluation criteria to evolve simultaneously. The portfolio combines multiple views of progress, reducing dependence on any one LLM-generated objective.

%---------------------------------------
\subsection{Consensus Objective Aggregation}
\label{sec:consensus}
%---------------------------------------

\begin{figure}[!t]
\centering
\includegraphics[width=0.9\linewidth]{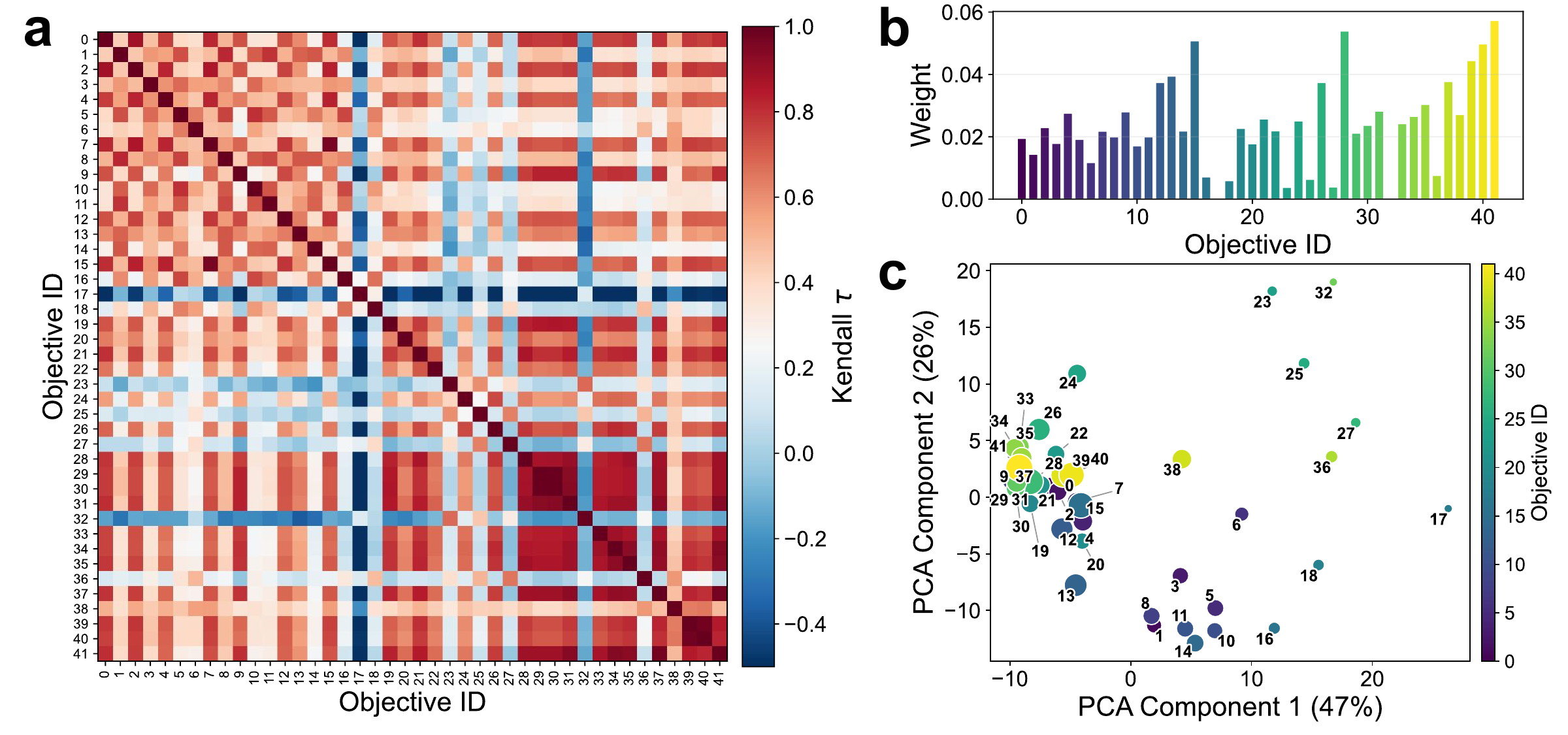}
\caption{\textbf{Consensus objective aggregation.} \textbf{(a)}~Kendall's~$\tau$ correlations across the 42 proxy objectives. Most objectives agree on design quality, while several outliers disagree with the majority. \textbf{(b)}~Raw consensus weights combine median agreement with recorded-round decay ($\lambda=0.9$). \textbf{(c)}~PCA of standardized objective rank vectors, colored by objective ID; larger dots carry more weight. Similar objectives cluster together, while dissenting objectives receive little weight.}
\label{fig:objective}
\end{figure}

No single proxy objective reliably captures solution quality. An objective rewarding low computational cost at small problem sizes, for example, may unintentionally favor solutions that scale poorly. Each time an LLM generates a new objective function, it can introduce subtle biases or overlook important edge cases. Depending on any one such objective, even one that evolves over time, can lead to reward hacking.

Instead, we maintain a portfolio of objective functions and aggregate them into a consensus ranking. The intuition is straightforward: each objective is an imperfect proxy for the same underlying research goal, and agreement across proxies provides a signal of shared support. The aggregation procedure (Fig.~\ref{fig:objective}) works as follows:

\begin{enumerate}
\item \textbf{Score matrix.} Each objective function $f_i$ scores every design $d_j$, producing a matrix $S_{ij} = f_i(d_j)$.
\item \textbf{Rank conversion.} Scores are converted to ranks within each objective (lower score = better rank), keeping the score range consistent across different objectives.
\item \textbf{Pairwise correlations.} We compute the Kendall's~$\tau$ rank correlation \citep{kendall_new_1938} $\tau_{ik}$ between every pair of objectives $f_i$ and $f_k$ over all designs. The complete Kendall's $\tau$ matrix is plotted in Fig.~\ref{fig:objective}(a), revealing clusters of positively correlated objectives and a few outliers.
\item \textbf{Agreement weighting.} Each objective's weight is proportional to its median pairwise $\tau$ with all other objectives (clamped at zero) multiplied by an exponential age decay: $w_i = \max(\widetilde{\tau}_i, 0) \cdot \lambda^{t - t_i}$, where $\widetilde{\tau}_i=\text{median}_{k\neq i}(\tau_{ik})$ is the median correlation, $\lambda = 0.9$ is the decay base, $t$ is the largest recorded creation-round index, and $t_i$ is the recorded creation-round index for objective $i$. Weights are then normalized to sum to one. The weights of all objectives are plotted in Fig.~\ref{fig:objective}(b).
\item \textbf{Consensus ranking.} The final ranking is a weighted Borda count \citep{emerson_original_2013}: $C_j = \sum_i w_i \cdot R_{ij} / (n - 1)$, where $R_{ij}$ is the rank of design $d_j$ under objective $f_i$ and $n$ is the number of designs.
\end{enumerate}

The agreement clamp gives zero weight to objectives with non-positive median correlation when other objectives retain positive weight. Age decay then modulates the relative influence of the remaining objectives. These operations are simple, inspectable, and independent of the numerical scales of the proxy scores.

Principal component analysis (PCA) on the standardized rank vectors of all objectives (Fig.~\ref{fig:objective}(c)) shows which objectives rank designs similarly. The portfolio evolved from power-law extrapolation toward metrics emphasizing the largest solved problem sizes. Appendix~\ref{sec:sm_objectives} traces this evolution and gives concrete examples of misleading proxies and their suppression.

Still, agreement-based weighting does have a blind spot. When a block of correlated objectives constitutes the majority, mutual agreement inflates every member's weight, even if the criterion they collectively encode is misaligned with the true research goal. The consensus mechanism cannot tell a genuinely informative majority from an echo chamber of redundant proxies.

To address this, the meta-agent periodically reviews all objectives and assigns weight multipliers $m_i$ that modulate the consensus weights ($w_i' \propto w_i \cdot m_i$). This gives the system a second way to revise its evaluation criteria, using context beyond rank agreement. In the 3-SAT case study, the meta-agent progressively suppressed early extrapolation objectives and emphasized large-$N$ performance. The resulting ranking is stable under changes to tie handling and retains nine of its top ten designs when the meta-agent multipliers are applied (Sec.~\ref{sec:repeated_runs} and Appendix~\ref{sec:implementation}).

%---------------------------------------
\subsection{Exploration via Monte Carlo Graph Search}
\label{sec:mcgs}
%---------------------------------------

The exploration-exploitation dilemma \citep{berger-tal_explorationexploitation_2014} is a well-studied problem and has standard solutions in bandit-style problems \citep{auer_finitetime_2002}. In our framework, we balance the exploration of novel designs and exploitation of promising designs via Monte Carlo Graph Search (MCGS)~\citep{czech_improving_2021}. Each new design is built from one or more reference designs, and this parent-child structure forms a directed graph, which is the basis for MCGS.

We adapt the Upper Confidence Bound (UCB) algorithm~\citep{auer_finitetime_2002,silver_mastering_2016,silver_mastering_2017} to our design graph. Each design node $j$ is scored by
\begin{equation}
\text{UCB}_j = \underbrace{r_j}_{\text{exploitation}} + \underbrace{c \cdot \frac{\sqrt{N_{\text{total}}}}{1 + n_j}}_{\text{exploration}},
\label{eq:ucb}
\end{equation}
where $r_j \in [0, 1]$ is the normalized rank score from the consensus objective (higher is better), $n_j$ is the visit count for design $j$, $N_{\text{total}}$ is the sum of all visit counts, and $c = 0.1$ is the exploration constant.

Visit counts measure how thoroughly a lineage has been explored. Our definition is different from standard MCTS~\citep{silver_mastering_2016,silver_mastering_2017} to accommodate the graph structure, variable parent influences, and the fact that any node can be expanded. When a design is created, the designer agent self-reports how much each parent contributed (weights summing to 1). These weighted counts propagate upward through the genealogy by breadth-first search with a depth-increasing decay ($\kappa = 0.9$), concentrating credit near the immediate parents. Designs referenced often as parents accumulate high visit counts, shrinking their exploration bonus and moving the planner toward less-explored but potentially rewarding parts of the design space.

As the graph is rebuilt every iteration from the current consensus objective, rankings always reflect the latest evaluation criteria. The planner agent picks high-UCB designs as starting points for new research directions.

%---------------------------------------
\subsection{Application: Algorithm Discovery for Combinatorial Optimization}
\label{sec:application}
%---------------------------------------

\begin{figure}[!t]
\centering
\includegraphics[width=\linewidth]{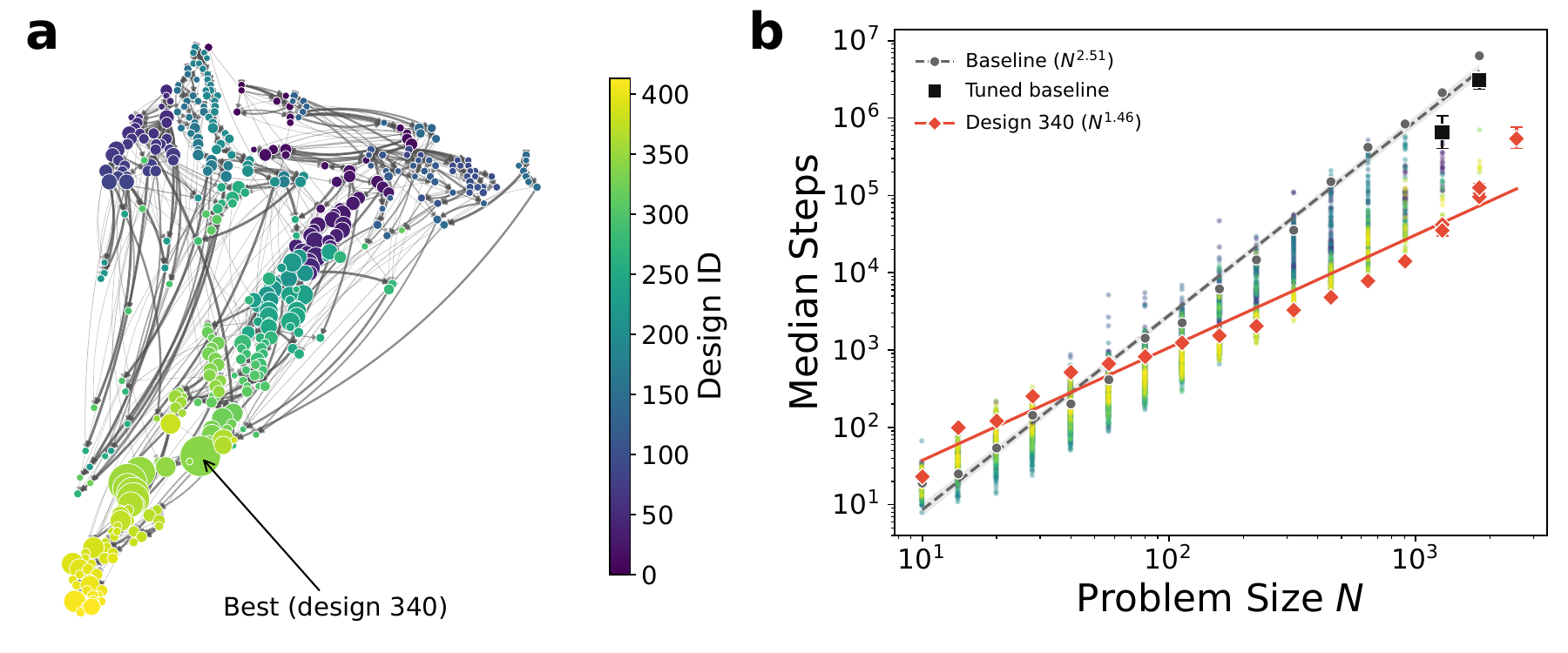}
\caption{\textbf{Algorithm discovery for 3-SAT DMM solvers.} \textbf{(a)}~Design genealogy from the original search. Nodes are colored by design ID and sized by raw-consensus score; edges indicate parent references. Separate workspaces contribute distinct lineages to the merged search. \textbf{(b)}~Median solution steps versus problem size. The original baseline, HEBO-tuned baseline, and design~340 are highlighted against the search background. Design~340 combines the original and fresh-instance measurements, with a power-law fit $N^{1.46}$ to all its points; the baseline retains its original $N^{2.51}$ reference line. Error bars show 95\% confidence intervals for fresh-instance measurements, evaluated with the original memory bound $x_l\leq10^6$. The largest-size measurements bend above these guides, while the substantial performance advantage persists.}
\label{fig:design}
\label{fig:benchmark}
\end{figure}

We test our framework on a challenging problem: designing efficient solvers for planted random 3-SAT instances (instances from Ref.~\citep{barthel_hiding_2002}) at clause-to-variable ratio 4.3 and planting parameter $p_0=0.08$. The baseline digital MemComputing machine (DMM) solver~\citep{bearden_efficient_2020,diventra_memcomputing_2022} is a physics-inspired dynamical system, whose fixed points encode satisfying assignments of the Boolean formula (see Sec.~\ref{sec:methods_DMM}). A key challenge is that solver performance at certain sizes does not necessarily predict scaling: a solver that looks fast at $N \lesssim 100$ variables may scale poorly to $N > 1000$, while benchmarking at large $N$ can be prohibitively expensive. This makes the choice of the evaluation criteria especially important.

Our system operated across several individual workspaces, producing a combined archive of 414 solver designs evaluated by 42 objectives. Selected source designs were imported with their genealogies, allowing the planner to draw on several exploratory lineages (Appendix~\ref{sec:sm_workspace_merging}). Over the course of the search, objectives grew steadily more sophisticated, from simple power-law fits early on to schedule-aware reach metrics that emphasize the largest solvable problem size, penalize budget cliffs, and analyze tail scaling.

Fig.~\ref{fig:design}(a) shows how the search builds on earlier designs, while Fig.~\ref{fig:design}(b) tracks the improvement in large-instance performance. Design~340, the leading design under the original raw-consensus ranking, adds a targeted release mechanism to the short-term memory dynamics. Appendix~\ref{sec:sm_design340} gives its architecture and genealogy. The following benchmark (Sec.~\ref{sec:benchmark}) evaluates this solver on fresh instances, and the objective ablation examines what changes when the evaluation criterion is held fixed.

The multi-fidelity schedule concentrates evaluation on promising candidates. Every design was first evaluated at low fidelity to filter out poor candidates quickly. Fidelity promotion is rule-based: designs with normalized consensus scores below 0.5 advance to medium fidelity, and those scoring below 0.1 advance to high fidelity (Appendix~\ref{sec:sm_schedule}).

%%%%%%%%%%%%%%%%%%%%%%%%%%%%%%%%%%%%%%%%%%%%%%%%%%%%%%%%%%%%%%%%%%%%%%%%%%%%%%%
\subsection{Performance at Larger Problem Sizes}
\label{sec:benchmark}

Design~340 retains its advantage when evaluated on fresh instances against a separately HEBO-tuned baseline (Table~\ref{tab:benchmark}). At $N=1810$, it requires 125,346 median integration steps compared with 3,096,796 for the tuned baseline, a $24.7\times$ reduction. At $N=2560$, with the same enlarged memory bound applied to both solvers (Sec.~\ref{sec:methods_DMM}), design~340 requires 654,967 median steps while the baseline has not reached its median after 25 million steps: a greater-than-$38\times$ reduction. The benefit therefore extends beyond the sizes reached during the original search.

\begin{table}[htbp]
\centering
\caption{\textbf{Design~340 versus a tuned baseline on fresh planted 3-SAT instances.} Median integration steps in thousands, with 95\% confidence intervals. The final baseline entry is a lower bound because fewer than half of its instances solve within 25 million steps.}
\label{tab:benchmark}
\small
\begin{tabular}{lccc}
\toprule
Solver & $N=1280$ & $N=1810$ & $N=2560$ \\
\midrule
Tuned baseline & 655 [405,1066] & 3097 [2344,3408] & $>25{,}000$ \\
Design 340 & 35 [30,40] & 125 [106,142] & 655 [433,868] \\
\bottomrule
\end{tabular}
\end{table}

Fig.~\ref{fig:design}(b) shows the scaling of both the baseline solver and the discovered design~340. A power-law fit is plotted for both solvers; at the largest sizes the curves turn upward, with design~340 still requiring far fewer steps than the baseline. Appendix~\ref{sec:protocol} summarizes the benchmark conditions.

\subsection{Objective Ablation: Fixed Expertise and Evolving Consensus}
\label{sec:repeated_runs}

To assess the contribution of consensus objective aggregation, we perform additional experiments with different objective configurations. The key ablation holds the initial expert objective fixed and removes objective evolution and aggregation. This expert objective is objective~0 in the consensus workspace: it blends power-law and exponential fits to predict solution cost at $N=2000$. A second ablation removes meta-agent guidance and weight multipliers while retaining objective generation, correlation weighting, and age decay. We also test fixed objectives selected to expose saturation and ranking misalignment. All configurations use the same planner/designer engine and periodic HEBO tuning, starting from the same inherited solver.

We compare progress using \emph{reach}: the largest tested problem size at which more than half of the 100 instances solve within the prescribed search budget. This is a manually chosen evaluation criterion, grounded in the stated research goal of solving larger instances efficiently. The majority-solved threshold corresponds to median solution time, and the schedule is held fixed across configurations. Reach measures demonstrated capability without fitting or extrapolating a scaling law; it does not require accepting any one search objective's predictions. Because it is discrete and does not distinguish solver speed within a size, we complement it with fresh-instance step counts below.

\begin{table}[htbp]
\centering
\caption{\textbf{Objective ablations.} The table measures \emph{reach}, the largest tested problem size at which more than half of the 100 instances solve within the prescribed search budget. All runs start at reach 453 (baseline solver). For consistency, we report reach after the first 56 new designs for every configuration.}
\label{tab:runs}
\small
\begin{tabular}{lrrr}
\toprule
Objective configuration & Reach @ 56 & Final reach & New designs \\
\midrule
Full consensus & 905 & 1280 & 166 \\
Consensus without meta-agent & 905 & 905 & 166 \\
Fixed expert objective 0, run 1 & 1280 & 2560 & 166 \\
Fixed expert objective 0, run 2 & 640 & 640 & 166 \\
Fixed solve rate at $N=320$ & 453 & 453 & 62 \\
Fixed LLM-written graded objective & 905 & 905 & 62 \\
Fixed average-log objective, run 1 & 453 & 453 & 62 \\
Fixed average-log objective, run 2 & 453 & 453 & 57 \\
\bottomrule
\end{tabular}
\end{table}

\textbf{Fixed expertise is effective, but its outcomes vary.} The first fixed-expert run reaches 2560 and produces design~89, which has the lowest median step count at $N=2560$ among the tested solvers (Table~\ref{tab:ablation_solvers}). The expert objective supplied to the consensus workspace thus provides useful guidance that the same search engine can exploit directly. Yet this success comes with substantial run-to-run variation: the second fixed-expert run, using the same search configuration, stops at 640. Full consensus reaches 1280 in the additional run and 1810 in the original campaign; the run without the meta-agent reaches 905.

\begin{table}[htbp]
\centering
\caption{\textbf{Solvers discovered under different objective configurations.} Median integration steps in thousands [95\% confidence interval]. Design~89 comes from the fixed-expert ablation. Designs~340 and~353 lead the original campaign's raw and meta-adjusted consensus rankings, respectively.}
\label{tab:ablation_solvers}
\small
\begin{tabular}{lcc}
\toprule
Discovered solver & $N=1810$ & $N=2560$ \\
\midrule
Fixed expert: design 89 & 68 [61,78] & 497 [327,665] \\
Consensus: design 340 & 125 [106,142] & 655 [433,868] \\
Consensus: design 353 & 113 [85,140] & 656 [536,807] \\
\bottomrule
\end{tabular}
\end{table}

\textbf{Consensus reduces exposure to a fixed proxy's blind spots.} Beyond this variation, the same expert objective can overrate solvers with limited performance evidence. In the original archive, objective~0 ranks design~247 second, above design~340 in sixth place, although design~247 was evaluated only through $N=320$ while design~340 demonstrated reach 1810. Design~247's nearly perfect small-size power-law fit ($R^2=0.9985$) effectively suppresses the competing exponential prediction. Raw consensus ranks this design 172nd. This exposes a reward-hacking vulnerability of the fixed proxy: a favorable small-size curve can earn a leading score before its large-instance performance has been demonstrated. Consensus discounts that shortcut by also rewarding observed performance at larger sizes.

\textbf{Consensus retains expert guidance within a stable ranking.} Consensus also preserves the expert objective's broader signal, retaining strong agreement with it across the archive (Spearman 0.961). Applying the last operative meta-agent multipliers also preserves nine of the top ten designs and gives a Spearman rank correlation of 0.976 with raw consensus. Together, these results show how consensus combines effective search guidance with reduced dependence on the blind spots of any single objective.

\section{Discussion}
%%%%%%%%%%%%%%%%%%%%%%%%%%%%%%%%%%%%%%%%%%%%%%%%%%%%%%%%%%%%%%%%%%%%%%%%%%%%%%%

We have demonstrated that treating scientific discovery as meta-optimization can produce substantial algorithmic improvements while automating the refinement of evaluation criteria. Consensus objective aggregation brings an adaptable portfolio of proxies into the research loop. In our 3-SAT DMM case study, the resulting solvers improve large-instance performance by more than an order of magnitude over a tuned baseline, with a greater-than-$38\times$ reduction in median integration steps at $N=2560$.

The practical value of the framework is that it develops solver code and refines how progress is measured within the same loop. Early power-law objectives gave way to metrics grounded in observed large-$N$ performance. The objective ablation shows complementary strengths: a fixed expert objective produced the fastest solver measured, while consensus provides an effective, stable ranking. A portfolio can retain the useful direction supplied by an expert proxy without letting that one proxy determine the entire search.

Agreement-based weighting can still reinforce shared biases, especially when all objectives use the same LLM backbone. More diverse objective generation is a natural next step. Our case study covers one backbone and a planted 3-SAT family; the observed fixed-objective run spread motivates a larger study of search variability.

There is also an inherent tension in how much autonomy the system should have. We opted for a structured design: fixed agent roles, predefined workflow stages, and guardrails constraining what each agent can touch. This constrains how experiments can be changed, though it is less flexible than agentic systems with full authorization in a sandbox environment. Our preliminary experiments with fewer restrictions revealed a few failure modes, including emergent goal drift, ``cheating'' by modifying evaluation scripts, and cherry-picking favorable results. How to balance the safety of structured workflows against the creative reach of unconstrained agents remains an important challenge.

%%%%%%%%%%%%%%%%%%%%%%%%%%%%%%%%%%%%%%%%%%%%%%%%%%%%%%%%%%%%%%%%%%%%%%%%%%%%%%%
\section{Methods}
\label{sec:methods}
%%%%%%%%%%%%%%%%%%%%%%%%%%%%%%%%%%%%%%%%%%%%%%%%%%%%%%%%%%%%%%%%%%%%%%%%%%%%%%%
% Target: ~1,000-1,500 words

%---------------------------------------
\subsection{Agent Architecture}
\label{sec:architecture}
%---------------------------------------

All four agents run on GPT-5.2~\citep{gpt5.2}, with structured outputs enforced through Pydantic schemas for reliable parsing. The planner takes in the MCGS-ranked design list together with experiment summaries (genealogies, performance metrics, failure patterns) and returns strategic research directions, each specifying reference design IDs and a natural-language description of the intended modification. The designer receives one direction, writes Python solver code, and launches multi-fidelity experiments. Every ten planner iterations (20 LLM proposals), the meta-agent analyzes objective agreement, assigns weight multipliers, identifies the current research phase, and issues strategic guidance to the objective agent. The objective agent produces a Python function mapping experiment results to a scalar score. A HEBO~\citep{cowen-rivers_hebo_2022} module runs 50 trials on the highest-MCGS-scoring untuned design when the fraction of tuned designs falls below 10\%, refining continuous hyperparameters without changing the solver architecture. Appendices~\ref{sec:sm_methodology}--\ref{sec:sm_prompts} describe the workflow, modular solver framework, and prompt templates. The supplied code and records document the implemented search.

%---------------------------------------
\subsection{Consensus Objective Algorithm}
\label{sec:consensus_methods}
%---------------------------------------

Given $K$ objective functions $\{f_1, \ldots, f_K\}$ and $n$ designs $\{d_1, \cdots, d_n\}$ with experiment results, the consensus objective is computed as follows:

\begin{enumerate}
\item \textbf{Score matrix:} $S_{ij} = f_i(d_j)$ for each objective $f_i$ and design $d_j$. Failed evaluations, missing values, and NaNs are set to $+\infty$; returned infinities are retained.
\item \textbf{Rank matrix:} For each objective $i$, designs are sorted by $S_{ij}$ and assigned integer ranks $R_{ij} \in \{0, 1, \ldots, n-1\}$, breaking score ties by design ID as in the recorded implementation.
\item \textbf{Kendall's~$\tau$ matrix:} For each pair $(i, k)$, compute $\tau_{ik} = \text{KendallTau}(\{R_{ij}\}_j, \{R_{kj}\}_j)$.
\item \textbf{Objective weights:}
\begin{equation}
w_i = \frac{\max\!\big(\widetilde{\tau}_i, 0\big) \cdot \lambda^{t - t_i}}{\sum_{k=1}^{K} \max\!\big(\widetilde{\tau}_k, 0\big) \cdot \lambda^{t - t_k}},
\label{eq:weights}
\end{equation}
where $\widetilde{\tau}_i = \text{median}_{k \neq i}\, \tau_{ik}$ is the median pairwise correlation, $\lambda = 0.9$ is the age decay base, $t=\max_i t_i$, and $t_i$ is the recorded creation-round index of objective $i$. If all unnormalized weights vanish, the implementation uses uniform weights; a single objective receives weight one.
\item \textbf{Meta-agent adjustment (optional):} If meta-agent multipliers $\{m_i\}$ are provided, weights are updated as $w_i' = w_i \cdot m_i / \sum_k w_k \cdot m_k$, with uniform weights if the denominator vanishes.
\item \textbf{Consensus score:}
\begin{equation}
C_j = \sum_{i=1}^{K} w_i \cdot \frac{R_{ij}}{n - 1},
\label{eq:consensus}
\end{equation}
where $C_j \in [0, 1]$ with lower values indicating better designs. For the meta-adjusted variant, replace $w_i$ by $w_i'$. This is a weighted Borda count~\citep{emerson_original_2013} that aggregates normalized ranks across all weighted objectives.
\end{enumerate}

For $n=1$, the normalized rank is zero. A tie-aware sensitivity analysis preserves the winner and top-10 set (Appendix~\ref{sec:implementation}). For designs evaluated on-the-fly during the search and not yet stored in the database, normalized ranks are estimated by binary search into each objective's sorted score distribution.

%---------------------------------------
\subsection{Monte Carlo Graph Search}
%---------------------------------------

We adapt Monte Carlo Graph Search (MCGS)~\citep{czech_improving_2021} from board games to the design space. Each design is a node in a directed acyclic graph whose edges are defined by the reference weights in the design's genealogy: when the designer creates design $d_j$ from parents $p_1, \ldots, p_m$ with weights $\omega_1, \ldots, \omega_m$ (self-reported, normalized to sum to 1), edges $(p_k \to d_j)$ are added for every $k \in [1, ..., m]$.

Each node starts with one visit. Visit counts $n_j$ for each design $d_j$ propagate upward by breadth-first search. Creating design $d_j$ with reference weight $\omega_k$ to parent $p_k$ adds $\omega_k$ to $p_k$'s visit count $n_k$. Propagation continues recursively, each hop from depth $d$ to $d{+}1$ attenuated by the ancestor's edge weight times $\kappa^{d+1}$ ($\kappa = 0.9$); deeper ancestors therefore receive rapidly diminishing credit. Propagation halts once contributions drop below $10^{-4}$.

The UCB score (Eq.~\eqref{eq:ucb}) combines the exploitation term $r_j$ (normalized consensus rank; the best design scores 1.0) and the exploration term ($c \cdot \frac{\sqrt{N_{\text{total}}}}{1 + n_j}$), with $N_{\text{total}}=\sum_j n_j$ and $c = 0.1$ setting the balance. Rebuilding the graph each iteration from the current consensus objective keeps rankings updated with the latest evaluation criteria.

%---------------------------------------
\subsection{Multi-Fidelity Schedule}
%---------------------------------------

Computational resources are allocated progressively across three fidelity tiers. Problem sizes follow a geometric progression $N_k = \mathrm{round}(N_0 \cdot 2^{k/2})$ with $N_0 = 10$ and $k = 0, 1, 2, \ldots$, yielding the sequence $N \in \{10, 14, 20, 28, 40, \ldots\}$. The solver step budget (\texttt{max\_steps}) starts at 50 and doubles at each level.

\begin{itemize}
\item \textbf{Low fidelity:} $N \leq 640$, \texttt{max\_steps} up to $10^4$, nominal timeout 1 minute. Fast initial screening.
\item \textbf{Medium fidelity:} $N \leq 1280$, \texttt{max\_steps} up to $10^5$, nominal timeout 5 minutes. Moderate evaluation for promising candidates.
\item \textbf{High fidelity:} $N \leq 2560$, \texttt{max\_steps} up to $10^6$, nominal timeout 30 minutes. Thorough benchmark of the best designs.
\end{itemize}

At every level, the solver is run on 100 random planted 3-SAT instances \citep{barthel_hiding_2002} per problem size at clause-to-variable ratio $\alpha_r = 4.3$ and $p_0=0.08$. A problem size is considered solved if more than half of the instances finish within the step budget. Fidelity promotion uses normalized consensus-score thresholds of 0.5 for medium fidelity and 0.1 for high fidelity. These thresholds act on averaged ranks and need not select exactly 50\% or 10\% of the population. The schedule stops a batch once the unsolved fraction drops below 0.5, so successful search records commonly report 0.49. These records guide search; the fresh-instance benchmark instead records per-instance first-solve times and handles censoring explicitly (Appendix~\ref{sec:protocol}).

%---------------------------------------
\subsection{Baseline 3-SAT DMM Algorithm}
\label{sec:methods_DMM}
%---------------------------------------

A random 3-SAT instance has $N$ Boolean variables $V_i \in \{0, 1\}$ and $M = \lfloor \alpha_r N \rfloor$ clauses, each being the disjunction of three literals. The baseline DMM algorithm relaxes every variable to a continuous value $v_i \in [-1, 1]$ and introduces auxiliary memory variables: a long-term clause penalty weight $x_{l,m} \in [1, 10^6]$ (enlarged to $x_{l,m} \in [1, 10^4M]$ for the tabulated comparisons at $N=2560$~\citep{bearden_efficient_2020}) that grows for persistently violated clauses, and a short-term satisfaction switch $x_{s,m} \in [0, 1]$ toggling between push and hold modes. The coupled ordinary differential equations governing the dynamics are:
\begin{align}
\dot{v}_n &= \sum_{m=1}^{M} \bigl[x_{l,m}\, x_{s,m}\, G_{n,m} + (1 + \zeta\, x_{l,m})(1 - x_{s,m})\, R_{n,m}\bigr], \label{eq:dmm_v} \\
\dot{x}_{s,m} &= \beta\, (x_{s,m} + \epsilon)\, (c_m - \gamma), \label{eq:dmm_xs} \\
\dot{x}_{l,m} &= \alpha\, (c_m - \delta), \label{eq:dmm_xl}
\end{align}

The literal polarity $q_{n,m} = \pm 1$ indicates whether variable $n$ appears positive or negated in clause $m$. The clause satisfaction monitor,
\begin{equation}
\begin{aligned}
&c_m(v_i, v_j, v_k) =\\
&\tfrac{1}{2}\min\bigl[(1 - q_{i,m}\, v_i),\; (1 - q_{j,m}\, v_j),\; (1 - q_{k,m}\, v_k)\bigr], \label{eq:dmm_c}
\end{aligned}
\end{equation}
ranges between 0 (satisfied) and 1 (violated). $\alpha, \beta, \gamma, \delta, \epsilon, \zeta$, and the integration step size $\Delta t_0$ are hyperparameters~\citep{bearden_efficient_2020}. The gradient term
\begin{equation}
G_{n,m} = \tfrac{1}{2}\, q_{n,m} \min\!\bigl[(1 - q_{j,m}\, v_j),\; (1 - q_{k,m}\, v_k)\bigr], \label{eq:dmm_G}
\end{equation}
pushes variable $v_n$ toward its satisfying polarity, scaled by how close the other two literals $j,k$ in the clause are to being satisfied. The rigidity term
\begin{equation}
R_{n,m} = \begin{cases}
\tfrac{1}{2}(q_{n,m} - v_n), & \text{if } c_m = \tfrac{1}{2}(1 - q_{n,m}\, v_n), \\
0, & \text{otherwise},
\end{cases} \label{eq:dmm_R}
\end{equation}
fires only when variable $n$ is the most satisfied literal in clause $m$, pulling it back toward its satisfying value to keep the clause from flipping. Both $G_{n,m}$ and $R_{n,m}$ are zero when variable $n$ is absent from clause $m$.

Finally, the numerical integration time step $\Delta t$ is rescaled by the inverse of the maximum absolute $v$-derivative:
\begin{equation}
\Delta t \;=\; \frac{1}{\max_{n} |\dot{v}_n|}\;\Delta t_0, \label{eq:dmm_scale}
\end{equation}
with the denominator lower-bounded at $10^{-6}$ for numerical stability. This normalized forward-Euler integrator is shared by the baseline and discovered solvers; it differs from the adaptive integration procedure in~\citep{bearden_efficient_2020}.

%%%%%%%%%%%%%%%%%%%%%%%%%%%%%%%%%%%%%%%%%%%%%%%%%%%%%%%%%%%%%%%%%%%%%%%%%%%%%%%

\ifdefined\arxivversion
\paragraph{Author contributions.}
Y.-H.Z. suggested and M.D. supervised the work. Y.-H.Z. designed and performed the numerical simulations. All authors contributed to the scientific discussion.

\paragraph{Code and data availability.}
The accompanying supplement separates a compact implementation of the framework from the supporting experimental data and optional logs. It includes the agent prompts, solver and objective definitions, measured results, and scripts for checking the reported analyses. The framework is also available at \url{https://github.com/yuanhangzhang98/LLM_meta_optimization}.

\paragraph{Acknowledgments.}
This work was funded by the National Science Foundation via grant No. ECCS-2229880. M.D. also acknowledges funding by the Alexander von Humboldt Stiftung through the Humboldt Research Award.

\else
\paragraph{Code and data availability.}
The anonymous supplement separates a compact implementation of the framework from the supporting experimental data and optional logs. It includes the agent prompts, solver and objective definitions, measured results, and scripts for checking the reported analyses.
\fi

\bibliographystyle{tmlr}
\bibliography{references}

@Book{OUPbook,
	Title                    = {The Scientific Method: Reflections from a Practitioner},
	Author                   = {{Di Ventra}, M.},
	Publisher                = {Oxford University Press, Oxford},
	Year                     = {2018}
}

@article{bearden_efficient_2020,
  title = {Efficient Solution of {{Boolean}} Satisfiability Problems with Digital Memcomputing},
  author = {Bearden, Sean R. B. and Pei, Yan Ru and Di Ventra, Massimiliano},
  year = 2020,
  month = nov,
  journal = {Scientific Reports},
  volume = {10},
  number = {1},
  pages = {19741},
  doi = {10.1038/s41598-020-76666-2},
  copyright = {2020 The Author(s)},
}

@article{boiko_autonomous_2023,
  title = {Autonomous Chemical Research with Large Language Models},
  author = {Boiko, Daniil A. and MacKnight, Robert and Kline, Ben and Gomes, Gabe},
  year = 2023,
  month = dec,
  journal = {Nature},
  volume = {624},
  number = {7992},
  pages = {570--578},
  doi = {10.1038/s41586-023-06792-0},
}

@article{czech_improving_2021,
  title = {Improving {{AlphaZero Using Monte-Carlo Graph Search}}},
  author = {Czech, Johannes and Korus, Patrick and Kersting, Kristian},
  year = 2021,
  month = may,
  journal = {Proceedings of the International Conference on Automated Planning and Scheduling},
  volume = {31},
  pages = {103--111},
  doi = {10.1609/icaps.v31i1.15952},
}

@misc{du_accelerating_2025,
  title = {Accelerating {{Scientific Discovery}} with {{Autonomous Goal-evolving Agents}}},
  author = {Du, Yuanqi and Yu, Botao and Liu, Tianyu and Shen, Tony and Chen, Junwu and Rittig, Jan G. and Sun, Kunyang and Zhang, Yikun and Song, Zhangde and Zhou, Bo and Masschelein, Cassandra and Wang, Yingze and Wang, Haorui and Jia, Haojun and Zhang, Chao and Zhao, Hongyu and Ester, Martin and {Head-Gordon}, Teresa and Gomes, Carla P. and Sun, Huan and Duan, Chenru and Schwaller, Philippe and Jin, Wengong},
  year = 2025,
  month = dec,
  number = {arXiv:2512.21782},
  eprint = {2512.21782},
  howpublished = {arXiv preprint arXiv:2512.21782},
  doi = {10.48550/arXiv.2512.21782},
}

@misc{eger_transforming_2025,
  title = {Transforming {{Science}} with {{Large Language Models}}: {{A Survey}} on {{AI-assisted Scientific Discovery}}, {{Experimentation}}, {{Content Generation}}, and {{Evaluation}}},
  author = {Eger, Steffen and Cao, Yong and D'Souza, Jennifer and Geiger, Andreas and Greisinger, Christian and Gross, Stephanie and Hou, Yufang and Krenn, Brigitte and Lauscher, Anne and Li, Yizhi and Lin, Chenghua and Moosavi, Nafise Sadat and Zhao, Wei and Miller, Tristan},
  year = 2025,
  month = apr,
  number = {arXiv:2502.05151},
  eprint = {2502.05151},
  howpublished = {arXiv preprint arXiv:2502.05151},
  doi = {10.48550/arXiv.2502.05151},
}

@article{emerson_original_2013,
  title = {The Original {{Borda}} Count and Partial Voting},
  author = {Emerson, Peter},
  year = 2013,
  month = feb,
  journal = {Social Choice and Welfare},
  volume = {40},
  number = {2},
  pages = {353--358},
  doi = {10.1007/s00355-011-0603-9},
}

@incollection{goodhart_problems_1984,
  title = {Problems of {{Monetary Management}}: {{The UK Experience}}},
  booktitle = {Monetary {{Theory}} and {{Practice}}: {{The UK Experience}}},
  author = {Goodhart, C. A. E.},
  editor = {Goodhart, C. A. E.},
  year = 1984,
  pages = {91--121},
  publisher = {Macmillan Education UK},
  address = {London},
  doi = {10.1007/978-1-349-17295-5_4},
}

@misc{gottweis_ai_2025,
  title = {Towards an {{AI}} Co-Scientist},
  author = {Gottweis, Juraj and Weng, Wei-Hung and Daryin, Alexander and Tu, Tao and Palepu, Anil and Sirkovic, Petar and Myaskovsky, Artiom and Weissenberger, Felix and Rong, Keran and Tanno, Ryutaro and Saab, Khaled and Popovici, Dan and Blum, Jacob and Zhang, Fan and Chou, Katherine and Hassidim, Avinatan and Gokturk, Burak and Vahdat, Amin and Kohli, Pushmeet and Matias, Yossi and Carroll, Andrew and Kulkarni, Kavita and Tomasev, Nenad and Guan, Yuan and Dhillon, Vikram and Vaishnav, Eeshit Dhaval and Lee, Byron and Costa, Tiago R. D. and Penad{\'e}s, Jos{\'e} R. and Peltz, Gary and Xu, Yunhan and Pawlosky, Annalisa and Karthikesalingam, Alan and Natarajan, Vivek},
  year = 2025,
  month = feb,
  number = {arXiv:2502.18864v1},
  eprint = {2502.18864v1},
  howpublished = {arXiv preprint arXiv:2502.18864v1},
  url = {https://arxiv.org/abs/2502.18864v1},
}

@misc{han_physagent_2025,
  title = {{{PhysAgent}}: {{A Multi-Agent Approach}} to the {{Automated Discovery}} of {{Physical Laws}}},
  author = {Han, Xiao-Qi and Gao, Ze-Feng and Guo, Peng-Jie and Lu, Zhong-Yi},
  year = 2025,
  month = aug,
  howpublished = {Qeios},
  doi = {10.32388/J2MXUW},
  copyright = {http://creativecommons.org/licenses/by/4.0/},
}

@misc{lu_ai_2024,
  title = {The {{AI Scientist}}: {{Towards Fully Automated Open-Ended Scientific Discovery}}},
  author = {Lu, Chris and Lu, Cong and Lange, Robert Tjarko and Foerster, Jakob and Clune, Jeff and Ha, David},
  year = 2024,
  month = sep,
  number = {arXiv:2408.06292},
  eprint = {2408.06292},
  howpublished = {arXiv preprint arXiv:2408.06292},
  doi = {10.48550/arXiv.2408.06292},
}

@article{lu_discovery_2025,
  title = {Discovery of the Reward Function for Embodied Reinforcement Learning Agents},
  author = {Lu, Renzhi and Shao, Zonghe and Ding, Yuemin and Chen, Ruijuan and Wu, Dongrui and Su, Housheng and Yang, Tao and Zhang, Fumin and Wang, Jun and Shi, Yang and Jiang, Zhong-Ping and Ding, Han and Zhang, Hai-Tao},
  year = 2025,
  month = dec,
  journal = {Nature Communications},
  volume = {16},
  number = {1},
  pages = {11064},
  doi = {10.1038/s41467-025-66009-y},
}

@misc{luo_llm4sr_2025,
  title = {{{LLM4SR}}: {{A Survey}} on {{Large Language Models}} for {{Scientific Research}}},
  author = {Luo, Ziming and Yang, Zonglin and Xu, Zexin and Yang, Wei and Du, Xinya},
  year = 2025,
  month = jan,
  number = {arXiv:2501.04306},
  eprint = {2501.04306},
  howpublished = {arXiv preprint arXiv:2501.04306},
  doi = {10.48550/arXiv.2501.04306},
}

@article{merchant_scaling_2023,
  title = {Scaling Deep Learning for Materials Discovery},
  author = {Merchant, Amil and Batzner, Simon and Schoenholz, Samuel S. and Aykol, Muratahan and Cheon, Gowoon and Cubuk, Ekin Dogus},
  year = 2023,
  month = dec,
  journal = {Nature},
  volume = {624},
  number = {7990},
  pages = {80--85},
  doi = {10.1038/s41586-023-06735-9},
}

@misc{mitchener_kosmos_2025,
  title = {Kosmos: {{An AI Scientist}} for {{Autonomous Discovery}}},
  author = {Mitchener, Ludovico and Yiu, Angela and Chang, Benjamin and Bourdenx, Mathieu and Nadolski, Tyler and Sulovari, Arvis and Landsness, Eric C. and Barabasi, Daniel L. and Narayanan, Siddharth and Evans, Nicky and Reddy, Shriya and Foiani, Martha and Kamal, Aizad and Shriver, Leah P. and Cao, Fang and Wassie, Asmamaw T. and Laurent, Jon M. and {Melville-Green}, Edwin and Caldas, Mayk and Bou, Albert and Roberts, Kaleigh F. and Zagorac, Sladjana and Orr, Timothy C. and Orr, Miranda E. and Zwezdaryk, Kevin J. and Ghareeb, Ali E. and McCoy, Laurie and Gomes, Bruna and Ashley, Euan A. and Duff, Karen E. and Buonassisi, Tonio and Rainforth, Tom and Bateman, Randall J. and Skarlinski, Michael and Rodriques, Samuel G. and Hinks, Michaela M. and White, Andrew D.},
  year = 2025,
  month = nov,
  number = {arXiv:2511.02824},
  eprint = {2511.02824},
  howpublished = {arXiv preprint arXiv:2511.02824},
  doi = {10.48550/arXiv.2511.02824},
}

@misc{novikov_alphaevolve_2025,
  title = {{{AlphaEvolve}}: {{A}} Coding Agent for Scientific and Algorithmic Discovery},
  author = {Novikov, Alexander and Vũ, Ngân and Eisenberger, Marvin and Dupont, Emilien and Huang, Po-Sen and Wagner, Adam Zsolt and Shirobokov, Sergey and Kozlovskii, Borislav and Ruiz, Francisco J. R. and Mehrabian, Abbas and Kumar, M. Pawan and See, Abigail and Chaudhuri, Swarat and Holland, George and Davies, Alex and Nowozin, Sebastian and Kohli, Pushmeet and Balog, Matej},
  year = 2025,
  month = jun,
  number = {arXiv:2506.13131},
  eprint = {2506.13131},
  howpublished = {arXiv preprint arXiv:2506.13131},
  doi = {10.48550/arXiv.2506.13131},
}

@article{peherstorfer_survey_2018,
  title = {Survey of {{Multifidelity Methods}} in {{Uncertainty Propagation}}, {{Inference}}, and {{Optimization}}},
  author = {Peherstorfer, Benjamin and Willcox, Karen and Gunzburger, Max},
  year = 2018,
  month = jan,
  journal = {SIAM Review},
  volume = {60},
  number = {3},
  pages = {550--591},
  doi = {10.1137/16M1082469},
}

@article{romera-paredes_mathematical_2024,
  title = {Mathematical Discoveries from Program Search with Large Language Models},
  author = {{Romera-Paredes}, Bernardino and Barekatain, Mohammadamin and Novikov, Alexander and Balog, Matej and Kumar, M. Pawan and Dupont, Emilien and Ruiz, Francisco J. R. and Ellenberg, Jordan S. and Wang, Pengming and Fawzi, Omar and Kohli, Pushmeet and Fawzi, Alhussein},
  year = 2024,
  month = jan,
  journal = {Nature},
  volume = {625},
  number = {7995},
  pages = {468--475},
  doi = {10.1038/s41586-023-06924-6},
  copyright = {2023 The Author(s)},
}

@article{silver_mastering_2016,
  title = {Mastering the Game of {{Go}} with Deep Neural Networks and Tree Search},
  author = {Silver, David and Huang, Aja and Maddison, Chris J. and Guez, Arthur and Sifre, Laurent and Van Den Driessche, George and Schrittwieser, Julian and Antonoglou, Ioannis and Panneershelvam, Veda and Lanctot, Marc and Dieleman, Sander and Grewe, Dominik and Nham, John and Kalchbrenner, Nal and Sutskever, Ilya and Lillicrap, Timothy and Leach, Madeleine and Kavukcuoglu, Koray and Graepel, Thore and Hassabis, Demis},
  year = 2016,
  month = jan,
  journal = {Nature},
  volume = {529},
  number = {7587},
  pages = {484--489},
  doi = {10.1038/nature16961},
}

@article{silver_mastering_2017,
  title = {Mastering the Game of {{Go}} without Human Knowledge},
  author = {Silver, David and Schrittwieser, Julian and Simonyan, Karen and Antonoglou, Ioannis and Huang, Aja and Guez, Arthur and Hubert, Thomas and Baker, Lucas and Lai, Matthew and Bolton, Adrian and Chen, Yutian and Lillicrap, Timothy and Hui, Fan and Sifre, Laurent and Van Den Driessche, George and Graepel, Thore and Hassabis, Demis},
  year = 2017,
  month = oct,
  journal = {Nature},
  volume = {550},
  number = {7676},
  pages = {354--359},
  doi = {10.1038/nature24270},
}

@article{sipling_phasespace_2026,
  title = {Phase-Space Engineering and Collective Dynamics in Memcomputing},
  author = {Sipling, Chesson and Zhang, Yuan-Hang and Di Ventra, Massimiliano},
  year = 2026,
  month = jan,
  journal = {Physical Review Applied},
  volume = {25},
  number = {1},
  pages = {014048},
  doi = {10.1103/f8tv-jv1b},
}

@article{skalse_defining_2022,
  title = {Defining and {{Characterizing Reward Gaming}}},
  author = {Skalse, Joar and Howe, Nikolaus and Krasheninnikov, Dmitrii and Krueger, David},
  year = 2022,
  month = dec,
  journal = {Advances in Neural Information Processing Systems},
  volume = {35},
  pages = {9460--9471},
}

@misc{su_end_2026,
  title = {The {{End}} of {{Reward Engineering}}: {{How LLMs Are Redefining Multi-Agent Coordination}}},
  author = {Su, Haoran and Sun, Yandong and Yu, Congjia},
  year = 2026,
  month = jan,
  number = {arXiv:2601.08237},
  eprint = {2601.08237},
  howpublished = {arXiv preprint arXiv:2601.08237},
  doi = {10.48550/arXiv.2601.08237},
}

@misc{wang_automated_2025,
  title = {Automated {{Algorithmic Discovery}} for {{Scientific Computing}} through {{LLM-Guided Evolutionary Search}}: {{A Case Study}} in {{Gravitational-Wave Detection}}},
  author = {Wang, He and Zeng, Liang},
  year = 2025,
  month = nov,
  number = {arXiv:2508.03661},
  eprint = {2508.03661},
  howpublished = {arXiv preprint arXiv:2508.03661},
  doi = {10.48550/arXiv.2508.03661},
}

@misc{wang_planning_2025,
  title = {Planning of {{Heuristics}}: {{Strategic Planning}} on {{Large Language Models}} with {{Monte Carlo Tree Search}} for {{Automating Heuristic Optimization}}},
  author = {Wang, Hui and Zhang, Xufeng and Mu, Chaoxu},
  year = 2025,
  month = jun,
  number = {arXiv:2502.11422},
  eprint = {2502.11422},
  howpublished = {arXiv preprint arXiv:2502.11422},
  doi = {10.48550/arXiv.2502.11422},
}

@misc{wang_swarms_2025,
  title = {Swarms of {{Large Language Model Agents}} for {{Protein Sequence Design}} with {{Experimental Validation}}},
  author = {Wang, Fiona Y. and Lee, Di Sheng and Kaplan, David L. and Buehler, Markus J.},
  year = 2025,
  month = nov,
  number = {arXiv:2511.22311},
  eprint = {2511.22311},
  howpublished = {arXiv preprint arXiv:2511.22311},
  doi = {10.48550/arXiv.2511.22311},
}

@misc{wei_ai_2025,
  title = {From {{AI}} for {{Science}} to {{Agentic Science}}: {{A Survey}} on {{Autonomous Scientific Discovery}}},
  author = {Wei, Jiaqi and Yang, Yuejin and Zhang, Xiang and Chen, Yuhan and Zhuang, Xiang and Gao, Zhangyang and Zhou, Dongzhan and Wang, Guangshuai and Gao, Zhiqiang and Cao, Juntai and Qiu, Zijie and Hu, Ming and Ma, Chenglong and Tang, Shixiang and He, Junjun and Song, Chunfeng and He, Xuming and Zhang, Qiang and You, Chenyu and Zheng, Shuangjia and Ding, Ning and Ouyang, Wanli and Dong, Nanqing and Cheng, Yu and Sun, Siqi and Bai, Lei and Zhou, Bowen},
  year = 2025,
  month = oct,
  number = {arXiv:2508.14111},
  eprint = {2508.14111},
  howpublished = {arXiv preprint arXiv:2508.14111},
  doi = {10.48550/arXiv.2508.14111},
}

@misc{yamada_ai_2025,
  title = {The {{AI Scientist-v2}}: {{Workshop-Level Automated Scientific Discovery}} via {{Agentic Tree Search}}},
  author = {Yamada, Yutaro and Lange, Robert Tjarko and Lu, Cong and Hu, Shengran and Lu, Chris and Foerster, Jakob and Clune, Jeff and Ha, David},
  year = 2025,
  month = apr,
  number = {arXiv:2504.08066},
  eprint = {2504.08066},
  howpublished = {arXiv preprint arXiv:2504.08066},
  doi = {10.48550/arXiv.2504.08066},
}

@misc{yao_tree_2023,
  title = {Tree of {{Thoughts}}: {{Deliberate Problem Solving}} with {{Large Language Models}}},
  author = {Yao, Shunyu and Yu, Dian and Zhao, Jeffrey and Shafran, Izhak and Griffiths, Thomas L. and Cao, Yuan and Narasimhan, Karthik},
  year = 2023,
  month = dec,
  number = {arXiv:2305.10601},
  eprint = {2305.10601},
  howpublished = {arXiv preprint arXiv:2305.10601},
  doi = {10.48550/arXiv.2305.10601},
}

@misc{yu_autonomous_2025,
  title = {Autonomous {{Code Evolution Meets NP-Completeness}}},
  author = {Yu, Cunxi and Liang, Rongjian and Ho, Chia-Tung and Ren, Haoxing},
  year = 2025,
  month = sep,
  number = {arXiv:2509.07367},
  eprint = {2509.07367},
  howpublished = {arXiv preprint arXiv:2509.07367},
  doi = {10.48550/arXiv.2509.07367},
}

@misc{zhang_accessing_2024,
  title = {Accessing {{GPT-4}} Level {{Mathematical Olympiad Solutions}} via {{Monte Carlo Tree Self-refine}} with {{LLaMa-3 8B}}},
  author = {Zhang, Di and Huang, Xiaoshui and Zhou, Dongzhan and Li, Yuqiang and Ouyang, Wanli},
  year = 2024,
  month = jun,
  number = {arXiv:2406.07394},
  eprint = {2406.07394},
  howpublished = {arXiv preprint arXiv:2406.07394},
  doi = {10.48550/arXiv.2406.07394},
}

@misc{zheng_automation_2025,
  title = {From {{Automation}} to {{Autonomy}}: {{A Survey}} on {{Large Language Models}} in {{Scientific Discovery}}},
  author = {Zheng, Tianshi and Deng, Zheye and Tsang, Hong Ting and Wang, Weiqi and Bai, Jiaxin and Wang, Zihao and Song, Yangqiu},
  year = 2025,
  month = sep,
  number = {arXiv:2505.13259},
  eprint = {2505.13259},
  howpublished = {arXiv preprint arXiv:2505.13259},
  doi = {10.48550/arXiv.2505.13259},
}

@misc{zheng_monte_2025,
  title = {Monte {{Carlo Tree Search}} for {{Comprehensive Exploration}} in {{LLM-Based Automatic Heuristic Design}}},
  author = {Zheng, Zhi and Xie, Zhuoliang and Wang, Zhenkun and Hooi, Bryan},
  year = 2025,
  month = jan,
  number = {arXiv:2501.08603},
  eprint = {2501.08603},
  howpublished = {arXiv preprint arXiv:2501.08603},
  doi = {10.48550/arXiv.2501.08603},
}

@article{kendall_new_1938,
  title = {A {{New Measure of Rank Correlation}}},
  author = {Kendall, M. G.},
  year = 1938,
  month = jun,
  journal = {Biometrika},
  volume = {30},
  number = {1-2},
  pages = {81--93},
  doi = {10.1093/biomet/30.1-2.81},
}

@article{barthel_hiding_2002,
  title = {Hiding {{Solutions}} in {{Random Satisfiability Problems}}: {{A Statistical Mechanics Approach}}},
  author = {Barthel, W. and Hartmann, A. K. and Leone, M. and {Ricci-Tersenghi}, F. and Weigt, M. and Zecchina, R.},
  year = 2002,
  month = apr,
  journal = {Physical Review Letters},
  volume = {88},
  number = {18},
  pages = {188701},
  doi = {10.1103/PhysRevLett.88.188701},
}

@misc{gpt5.2,
  title = {Introducing {{GPT-5}}.2},
  author = {{OpenAI}},
  year = 2025,
  month = dec,
  url = {https://openai.com/index/introducing-gpt-5-2/},
}

@article{cowen-rivers_hebo_2022,
  author = {Cowen-Rivers, Alexander and Lyu, Wenlong and Tutunov, Rasul and Wang, Zhi and Grosnit, Antoine and Griffiths, Ryan-Rhys and Maravel, Alexandre and Hao, Jianye and Wang, Jun and Peters, Jan and Bou Ammar, Haitham},
  year = {2022},
  month = {07},
  pages = {1269--1349},
  doi = {10.1613/jair.1.13643},
  title = {HEBO: Pushing The Limits of Sample-Efficient Hyperparameter Optimisation},
  volume = {74},
  journal = {Journal of Artificial Intelligence Research}
}

@book{diventra_memcomputing_2022,
  title = {{{MemComputing}}: {{Fundamentals}} and {{Applications}}},
  author = {Di Ventra, Massimiliano},
  year = 2022,
  month = feb,
  publisher = {Oxford University Press},
  doi = {10.1093/oso/9780192845320.001.0001},
}

@article{traversa_polynomialtime_2017,
  title = {Polynomial-Time Solution of Prime Factorization and {{NP-complete}} Problems with Digital Memcomputing Machines},
  author = {Traversa, Fabio L. and Di Ventra, Massimiliano},
  year = 2017,
  month = feb,
  journal = {Chaos: An Interdisciplinary Journal of Nonlinear Science},
  volume = {27},
  number = {2},
  pages = {023107},
  doi = {10.1063/1.4975761},
}

@article{auer_finitetime_2002,
  title = {Finite-Time {{Analysis}} of the {{Multiarmed Bandit Problem}}},
  author = {Auer, Peter and {Cesa-Bianchi}, Nicol{\`o} and Fischer, Paul},
  year = 2002,
  month = may,
  journal = {Machine Learning},
  volume = {47},
  number = {2},
  pages = {235--256},
  doi = {10.1023/A:1013689704352},
}

@article{berger-tal_explorationexploitation_2014,
  title = {The {{Exploration-Exploitation Dilemma}}: {{A Multidisciplinary Framework}}},
  author = {{Berger-Tal}, Oded and Nathan, Jonathan and Meron, Ehud and Saltz, David},
  year = 2014,
  month = apr,
  journal = {PLOS ONE},
  volume = {9},
  number = {4},
  pages = {e95693},
  doi = {10.1371/journal.pone.0095693},
}

@inproceedings{paszke_pytorch_2019,
  title = {{{PyTorch}}: {{An Imperative Style}}, {{High-Performance Deep Learning Library}}},
  booktitle = {Advances in {{Neural Information Processing Systems}}},
  author = {Paszke, Adam and Gross, Sam and Massa, Francisco and Lerer, Adam and Bradbury, James and Chanan, Gregory and Killeen, Trevor and Lin, Zeming and Gimelshein, Natalia and Antiga, Luca and Desmaison, Alban and Kopf, Andreas and Yang, Edward and DeVito, Zachary and Raison, Martin and Tejani, Alykhan and Chilamkurthy, Sasank and Steiner, Benoit and Fang, Lu and Bai, Junjie and Chintala, Soumith},
  year = 2019,
  volume = {32},
  publisher = {Curran Associates, Inc.},
}
\clearpage
\appendix
\section{Evaluation and Ablation Details}
\label{sec:protocol}

\subsection{Solver evaluation}
The benchmark uses planted random 3-SAT instances with clause-to-variable ratio 4.3 and planting parameter $p_0=0.08$, independently generated after the search. Solvers are evaluated using the same normalized forward-Euler integrator. The baseline's seven parameters, including the integration time step, are tuned with 50 HEBO trials. Discovered solvers receive the same number of trials per tuning round during architectural search. Median integration steps measure algorithmic work; they do not imply an equal wall-clock speedup.

Unsolved instances are retained as right-censored observations at their actual stopping step. We report Kaplan--Meier medians and batch-bootstrap 95\% intervals. Table~\ref{tab:benchmark} uses 500 instances for design~340 at each size and for the baseline at 1280, and 200 for the baseline at 1810 and 2560. Table~\ref{tab:ablation_solvers} uses 500 instances per entry except design~353 at 2560, which has 400. Standard evaluations allow up to four million steps; the baseline receives larger budgets of 20 million at 1810 and 25 million at 2560. Time limits can stop a run earlier, and unidentified medians remain censored rather than being extrapolated.

The original implementation uses the fixed long-term-memory bound $x_l\leq10^6$. At $N=2560$, the tabulated comparison applies the enlarged bound $x_l\leq10^4M$ from~\citet{bearden_efficient_2020} to every solver. Design~340's median is 540,985 steps with the fixed bound and 654,967 with the enlarged bound, so the gain persists under both conventions. With the enlarged bound, the baseline solves 79 of 200 instances within 25 million steps; the greater-than-$38\times$ comparison uses this censored median and design~340's point estimate.

At $N=3620$, the fixed-bound measurements for designs~89, 340, and 353 solve 22.3\%, 27.4\%, and 25.7\% of 1,000 instances, respectively; their medians remain unidentified. Another fixed-objective candidate, design~80, has median 1,733,311 steps at 2560. The supplementary data provide the complete measurements, implementation settings, and reproducibility information.

\subsection{Objective configurations}
The original archive combines selected lineages from several workspaces with a subsequent merged search. Its 414 retained designs therefore describe the discovery history rather than an equal-budget replicate of the additional runs. Each long ablation run contains 150 LLM proposals and 16 registered HEBO variants. The fixed-objective screens contain 56 proposals and six HEBO variants, except one average-log run that stopped at 52 proposals and five variants. The common checkpoint in Table~\ref{tab:runs} contains 51 proposals and five variants in every run. Different promotion decisions incur different evaluation costs, so equal design counts do not imply equal compute.

The solve-rate objective saturates once instances at $N=320$ are solved. The average-log objective can reward solvers with a low reach. The graded objective was written by an LLM and then held fixed. These objective types were selected after screening existing results for saturation and ranking misalignment, making them diagnostic probes of misspecification. The expert objective and the evaluation schedule were available to the objective and meta-agents; the evolving portfolio builds on this supplied guidance.

\subsection{Transfer beyond the planted family}
\label{sec:transfer}
On SAT-certified uniform unplanted 3-SAT at clause-to-variable ratio 4.25, the default baseline, tuned baseline, and design~340 all reach $N=640$. At that size, design~340 records 123,702 median steps versus 71,061 for the tuned baseline on 52 certified solvable instances. On the easier ratio-3.8 set, all three reach 2560, with recorded medians of 427 for design~340, 1,374 for the tuned baseline, and 3,065 for the default baseline. The learned modification's benefit thus depends on the instance distribution.

\section{Consensus Implementation}
\label{sec:implementation}

During the merged search, the implementation broke ties by design ID. Reanalysis with tie-aware Kendall $\tau_b$ and average-rank Borda scores preserves the winner and top-ten set, with rank correlation 0.99999.

Agreement does not eliminate shared bias. A 23-objective group connected by correlations of at least 0.7 carries 74\% of raw weight and 90\% after the last operative meta-agent multipliers. The meta-agent redirected influence toward larger-size performance, while this broader group remained dominant. Objective~32 illustrates the opposite case: its negative median agreement gives it zero raw weight. Consensus can therefore filter dissenting proxies, but whether the majority's criterion is useful still depends on the task and the evidence.

\section{Analysis of Design 340}
\label{sec:sm_design340}
%%%%%%%%%%%%%%%%%%%%%%%%%%%%%%%%%%%%%%%%%%%%%%%%%%%%%%%%%%%%%%%%%%%%%%%%%%%%%%%

Design~340 is the solver selected by the raw consensus ranking for the fresh-instance comparison in Fig.~\ref{fig:benchmark}. Below we trace its genealogy, compare its modifications to the baseline DMM equations (Eqs.~\eqref{eq:dmm_v}--\eqref{eq:dmm_xl}), and test which release components support its performance.

%---------------------------------------
\subsection{Architectural modifications}
\label{sec:sm_architecture}
%---------------------------------------

\begin{figure}[!t]
    \centering
    \includegraphics[width=0.95\linewidth]{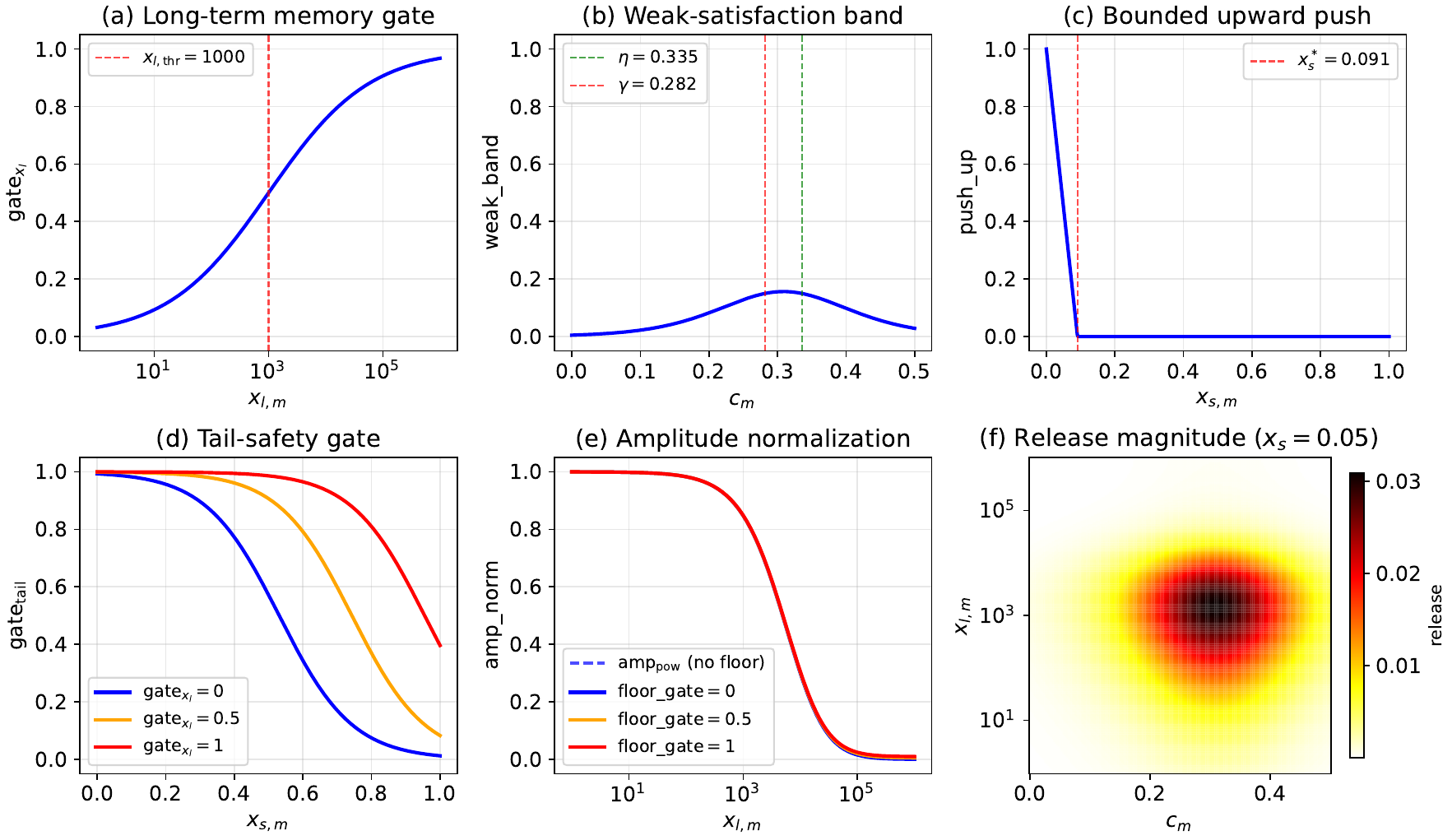}
    \caption{\textbf{Gating components of the release term in design~340.} \textbf{(a)}~Long-term memory gate $\mathrm{gate}_{x_l}$: reaches half-activation at $x_{l,\mathrm{thr}} \approx 1000$ (dashed red line). \textbf{(b)}~Weak-satisfaction band: peaks midway between $\eta$ and $\gamma$; note that the optimized values $\eta = 0.335 > \gamma = 0.282$ produce a narrow, low-amplitude response (peak $\approx 0.16$). \textbf{(c)}~Bounded upward push: linearly decreasing in $x_{s,m}$, reaching zero at $x_s^* = 0.091$. \textbf{(d)}~Tail-safety gate at three levels of $\mathrm{gate}_{x_l}$, showing how the $x_l$-dependent shift extends the safe operating range. \textbf{(e)}~Amplitude normalization with power-law decay and clause-state-driven floor at three levels of $\mathrm{floor\_gate}$. \textbf{(f)}~Combined release magnitude in the $(c_m, x_{l,m})$ plane at $x_{s,m} = 0.05$, showing the narrow region where all gates are simultaneously open.}
    \label{fig:sm_gating}
\end{figure}

Design~340 leaves the baseline variable dynamics~(Eq.~\eqref{eq:dmm_v}) and long-term memory dynamics~(Eq.~\eqref{eq:dmm_xl}) intact. The architectural modification is concentrated in the short-term memory equation~(Eq.~\eqref{eq:dmm_xs}), which acquires an additive release term:
\begin{equation}
    \dot{x}_{s,m} = \beta \bigl[\underbrace{(x_{s,m} + \epsilon)(c_m - \gamma)}_{\text{baseline}} + \underbrace{\mathcal{R}_m}_{\text{release}}\bigr],
    \label{eq:xs_340}
\end{equation}
where the release term $\mathcal{R}_m$ is a product of five factors that control when and how strongly the system intervenes:
\begin{equation}
    \mathcal{R}_m = \kappa \cdot \mathrm{gate}_{x_l} \cdot \mathrm{weak\_band} \cdot \mathrm{push\_up} \cdot \mathrm{gate}_{\mathrm{tail}} \cdot \mathrm{amp\_norm}.
    \label{eq:release}
\end{equation}
The multiplicative structure concentrates release where the factors overlap; the bounded push makes it exactly zero when $x_{s,m}\geq x_s^*$. Each component is described below and visualized in Fig.~\ref{fig:sm_gating}.

\textbf{1. Long-term memory gate} (Fig.~\ref{fig:sm_gating}(a)). A sigmoid in log-space that increases release for clauses carrying a large long-term memory:
\begin{equation}
    \mathrm{gate}_{x_l} = \sigma\!\left(\frac{\ln x_{l,m} - \ln x_{l,\mathrm{thr}}}{\omega_l}\right),
    \label{eq:gate_xl}
\end{equation}
with $x_{l,\mathrm{thr}} = 1000.4$ and $\omega_l = 2.03$. The threshold marks half-activation, favoring clauses with accumulated long-term memory.

\textbf{2. Weak-satisfaction band} (Fig.~\ref{fig:sm_gating}(b)). Two opposing sigmoids select clauses in a narrow satisfaction window:
\begin{equation}
    \mathrm{weak\_band} = \sigma\!\left(\frac{c_m - \eta}{\omega_b}\right) \cdot \sigma\!\left(\frac{\gamma - c_m}{\omega_b}\right),
    \label{eq:weak_band}
\end{equation}
where $\eta = 0.335$, $\gamma = 0.282$, and $\omega_b = 0.063$. Notably, the optimized values satisfy $\eta > \gamma$, yielding a very narrow, low-amplitude response (peak $\approx 0.16$). The band concentrates release around an intermediate clause cost $c_m\approx0.31$. The inverted ordering $\eta > \gamma$ emerged from hyperparameter optimization and makes the release mechanism highly selective, delivering only a small nudge to clauses in this narrow region.

\textbf{3. Bounded upward push} (Fig.~\ref{fig:sm_gating}(c)). A ReLU gate that drives $x_{s,m}$ toward a target value:
\begin{equation}
    \mathrm{push\_up} = \frac{\max(x_s^* - x_{s,m},\; 0)}{x_s^*},
    \label{eq:push_up}
\end{equation}
with $x_s^* = 0.091$. The target sits near the lower bound of $x_s$, so release shuts off once $x_s$ rises above $\sim 0.09$. Together with the weak-band, it concentrates the intervention in a small region of state space.

\textbf{4. Tail-safety gate} (Fig.~\ref{fig:sm_gating}(d)). A sigmoid that damps release when $x_{s,m}$ approaches its upper bound:
\begin{equation}
    \mathrm{gate}_{\mathrm{tail}} = \sigma\!\left(\frac{x_{s,\mathrm{tail}} - x_{s,m} + \mu_{\mathrm{tail}} \cdot \mathrm{gate}_{x_l}}{\omega_{\mathrm{tail}}}\right),
    \label{eq:gate_tail}
\end{equation}
with $x_{s,\mathrm{tail}} = 0.531$, $\mu_{\mathrm{tail}} = 0.424$, and $\omega_{\mathrm{tail}} = 0.107$. The $x_l$-dependent shift ($\mu_{\mathrm{tail}} \cdot \mathrm{gate}_{x_l}$) widens the safe operating range for high-penalty clauses, giving them more headroom before the safety cutoff kicks in. With the previous push\_up term concentrating on small $x_s$, this gate becomes almost redundant. The component ablation below tests its contribution at the optimized settings. 

\textbf{5. Amplitude normalization} (Fig.~\ref{fig:sm_gating}(e)). A power-law decay with a clause-state-driven floor:
\begin{align}
    \mathrm{amp}_{\mathrm{pow}} &= \left(\frac{x_{l,\mathrm{norm}}}{x_{l,\mathrm{norm}} + x_{l,m}}\right)^{\!p}, \label{eq:amp_pow} \\
    \mathrm{floor\_gate} &= 1 - (1 - \mathrm{weak\_band})(1 - \mathrm{push\_up}), \label{eq:floor_gate} \\
    \mathrm{amp\_norm} &= (1 - f)\,\mathrm{amp}_{\mathrm{pow}} + f, \quad f = a_{\mathrm{floor}} \cdot \mathrm{floor\_gate}, \label{eq:amp_norm}
\end{align}
where $x_{l,\mathrm{norm}} = 10{,}087$, $p = 1.75$, and $a_{\mathrm{floor}} = 0.0093$. The power-law decay (Eq.~\eqref{eq:amp_pow}) regulates the release term as $x_{l,m}$ increases. The floor gate (Eq.~\eqref{eq:floor_gate}) is a soft logical OR of the weak-band and push-up signals: it raises the amplitude floor when either signal is large. Hyperparameter optimization yields a small $a_{\mathrm{floor}}$, making this a modest correction to the power-law decay. 

Table~\ref{tab:hyperparams} gives the full hyperparameter comparison between the baseline and design~340.

\begin{table}[htbp]
    \centering
    \caption{\textbf{Hyperparameter comparison.} The original untuned baseline uses 7 parameters; design~340 uses 19, with the 12 additional parameters controlling the release mechanism. Values shown are the HEBO-optimized defaults for design~340. The performance comparison in Fig.~\ref{fig:benchmark} uses a separately tuned baseline.}
    \label{tab:hyperparams}
    \begin{tabular}{llcc}
        \toprule
        Parameter & Role & Baseline & Design 340 \\
        \midrule
        \multicolumn{4}{l}{\textit{Shared parameters (backbone)}} \\
        $\alpha$ & $x_l$ growth rate & 5.0 & 11.24 \\
        $\beta$ & $x_s$ update rate & 20.0 & 42.07 \\
        $\gamma$ & $x_s$ threshold & 0.25 & 0.282 \\
        $\delta$ & $x_l$ threshold & 0.05 & 0.080 \\
        $\epsilon$ & $x_s$ regularization & $10^{-3}$ & $8.5 \times 10^{-4}$ \\
        $\zeta$ & Rigidity coupling & $10^{-3}$ & 0.021 \\
        $\Delta t_0$ & Integration time step & 1.0 & 2.99 \\
        \midrule
        \multicolumn{4}{l}{\textit{Release mechanism (new)}} \\
        $\kappa$ & Release strength & --- & 1.015 \\
        $x_{l,\mathrm{thr}}$ & Memory gate threshold & --- & 1000.4 \\
        $\omega_l$ & Memory gate sharpness & --- & 2.025 \\
        $\eta$ & Band lower edge & --- & 0.335 \\
        $\omega_b$ & Band sharpness & --- & 0.063 \\
        $x_s^*$ & Push-up target & --- & 0.091 \\
        $x_{s,\mathrm{tail}}$ & Tail-safety threshold & --- & 0.531 \\
        $\omega_{\mathrm{tail}}$ & Tail-safety sharpness & --- & 0.107 \\
        $\mu_{\mathrm{tail}}$ & Tail $x_l$-shift & --- & 0.424 \\
        $x_{l,\mathrm{norm}}$ & Amp decay scale & --- & 10,087 \\
        $p$ & Amp decay power & --- & 1.753 \\
        $a_{\mathrm{floor}}$ & Amp floor level & --- & 0.0093 \\
        \bottomrule
    \end{tabular}
\end{table}

\textbf{Design choices and simplification.}
The five-factor structure was found without an explicit penalty for parameter count or code length. The wall-time limit constrains evaluation cost, while the component ablation below identifies opportunities to simplify the final release mechanism.

%---------------------------------------
\subsection{Genealogy}
\label{sec:sm_genealogy}
%---------------------------------------

\begin{figure}[!t]
    \centering
    \includegraphics[width=0.95\linewidth]{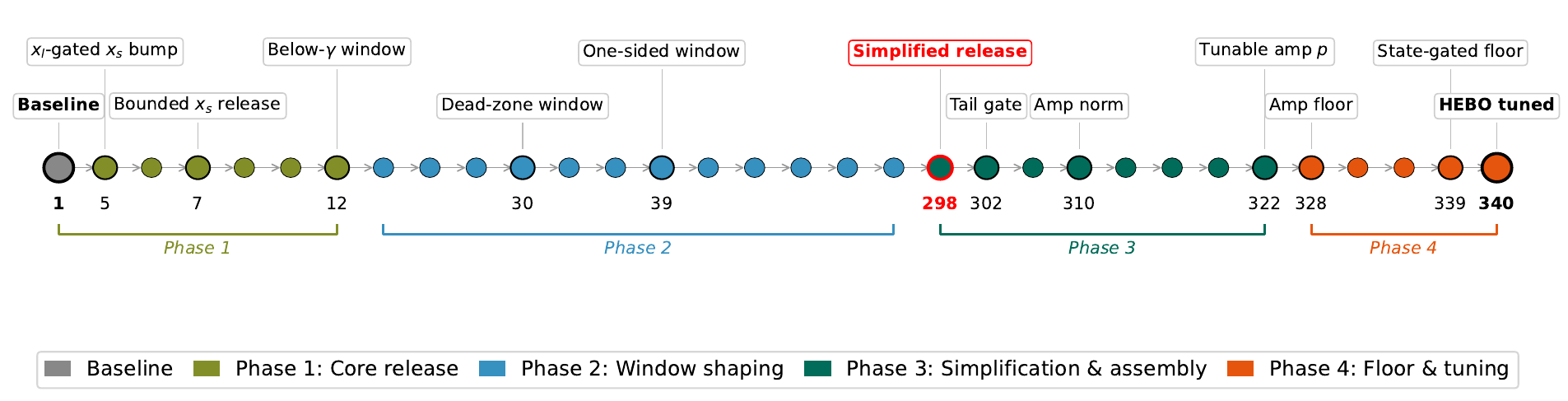}
    \caption{\textbf{Primary lineage of design~340.} Each node is a design on the 32-design primary ancestry path, colored by innovation phase. Labels identify selected milestones.}
    \label{fig:sm_design340_ancestry}
\end{figure}

Design~340 has a primary ancestry path containing 32 designs, including the baseline. Fig.~\ref{fig:sm_design340_ancestry} shows the primary lineage, traced by following the highest-weight parent reference at each step. Table~\ref{tab:lineage} lists every design in this lineage with its key innovation and maximum solved problem size. Along this path, the release mechanism is developed through changes to the short-term memory dynamics $\dot{x}_{s,m}$, while the variable and long-term memory equations retain their baseline form. Table~\ref{tab:lineage} reports historical discovery records; the fresh-instance benchmark is reported separately in Fig.~\ref{fig:benchmark}.

\newpage
\begingroup
\small
\setlength{\tabcolsep}{3pt}
\begin{longtable}{rp{3.0cm}p{5.2cm}rr}
\caption{\textbf{Primary lineage of design~340.} Historical discovery records along the highest-weight parent path. $N_{\max}$ is the largest recorded size with unsolved fraction below $0.5$. HEBO rows change parameter defaults while retaining the parent's dynamics.}\label{tab:lineage}\\
\toprule
ID & Short name & Innovation & $N_{\max}$ & Steps \\
\midrule
\endfirsthead
\multicolumn{5}{l}{\textit{Table \thetable\ continued}} \\
\toprule
ID & Short name & Innovation & $N_{\max}$ & Steps \\
\midrule
\endhead
        \multicolumn{5}{l}{\textit{Phase~1: Core release structure (designs 1--12)}} \\
1 & Baseline & --- & 320 & 37{,}729 \\
5 & $x_l$-gated $x_s$ bump & $\mathrm{gate}_{x_l}$, $\kappa$ & 320 & 9{,}241 \\
6 & $x_s$ hysteresis bump & $(1 - x_s)$ damping & 320 & 9{,}452 \\
7 & Bounded $x_s$ release & weak\_band, push\_up, $x_s^*$ & \textbf{453} & 66{,}681 \\
10 & Near-$\gamma$ release gate & near\_gamma gate & 453 & 45{,}234 \\
11 & HEBO-tuned & (hyperparameters) & 453 & 12{,}409 \\
12 & Below-$\gamma$ windowed & below-$\gamma$ filter & \textbf{905} & 229{,}470 \\
\midrule
        \multicolumn{5}{l}{\textit{Phase~2: Window shaping \& amplitude refinement (designs 16--285)}} \\
16 & Floored window & win\_floor & 905 & 203{,}389 \\
20 & Power-shaped window & win\_pow & 905 & 191{,}288 \\
21 & HEBO-tuned & (hyperparameters) & 640 & 32{,}071 \\
30 & Dead-zone window & win\_tau dead zone & \textbf{1{,}280} & 220{,}176 \\
34 & Smootherstep amp & smootherstep function & 1{,}280 & 230{,}042 \\
36 & Warp amp & rational warp & 1{,}280 & 225{,}071 \\
39 & One-sided window & simplified to one-sided & 1{,}280 & 130{,}867 \\
43 & $x_l$-gain amp & $x_l$-dependent boost & 1{,}280 & 191{,}946 \\
222 & Safe $x_l$-sharp cap & sharp\_min\_ratio, sharp\_gain & 905 & 35{,}104 \\
228 & $\gamma$-threshold shift & thr\_shift & 905 & 62{,}151 \\
280 & Shift sharpening & shift\_pow & 1{,}280 & 143{,}552 \\
285 & Dead-zone shift & shift\_tau & 320 & 3{,}476 \\
\midrule
        \multicolumn{5}{l}{\textit{Phase~3: Simplification \& component assembly (designs 298--322)}} \\
298 & Weak-band release & \textbf{radical simplification} & 640 & 64{,}272 \\
302 & Tail gate & gate\_tail, $x_{s,\mathrm{tail}}$ & 640 & 51{,}263 \\
305 & $x_l$-shifted tail & tail\_mu & 320 & 6{,}001 \\
310 & Amplitude norm & amp\_norm & 640 & 45{,}978 \\
316 & Sign flip in tail & $-\mu \to +\mu$ & 640 & 38{,}994 \\
320 & Over-damped & amp\_norm$^2$ & 640 & 58{,}343 \\
321 & HEBO-tuned & (hyperparameters) & \textbf{1{,}280} & 118{,}834 \\
322 & Tunable amp power & amp\_p & 905 & 28{,}192 \\
\midrule
        \multicolumn{5}{l}{\textit{Phase~4: Floor refinement \& final tuning (designs 328--340)}} \\
328 & $x_l$-gated floor & amp\_floor $\cdot$ gate$_{x_l}^2$ & 905 & 41{,}271 \\
332 & Linear floor & amp\_floor $\cdot$ gate$_{x_l}$ & 905 & 31{,}154 \\
336 & Ungated floor & amp\_floor (constant) & 320 & 2{,}916 \\
339 & State-gated floor & \textbf{floor\_gate} (soft-OR) & \textbf{1{,}810} & 691{,}058 \\
340 & HEBO-tuned & (hyperparameters) & \textbf{1{,}810} & \textbf{95{,}503} \\
\bottomrule
\end{longtable}
\endgroup

The evolution falls into four phases, each contributing distinct pieces of the final solver.

\subsubsection*{Phase~1: Core release structure (designs 1-12)}

The first phase laid down the release term's basic architecture. Design~5 introduced the central idea: an additive ``bump'' in the $x_s$ dynamics, gated by long-term memory: $\mathcal{R}_m = \kappa \cdot \mathrm{gate}_{x_l}(x_{l,m}) \cdot c_m \cdot \max(\gamma - c_m, 0)$. It fires for persistently violated clauses ($x_{l,m} \gg x_{l,\mathrm{thr}}$) whose satisfaction monitor falls below $\gamma$, cutting the baseline's median step count at $N=320$ from 37{,}729 to 9{,}241. Design~6 then added a $(1 - x_s)$ damping factor so the bump would not saturate $x_s$.

Design~7 established the core release structure. It replaced the ad-hoc $c_m \cdot \max(\gamma - c_m, 0) \cdot (1 - x_s)$ product with three cleanly separated components---$\mathrm{weak\_band}$, $\mathrm{push\_up}$, and $x_s^*$---establishing the multiplicative gating structure that survives into the final solver:
\begin{equation}
    \mathcal{R}_m = \kappa \cdot \mathrm{gate}_{x_l} \cdot \mathrm{weak\_band}(c_m) \cdot \mathrm{push\_up}(x_{s,m}).
    \label{eq:release_v7}
\end{equation}
$N_{\max}$ jumped from 320 to 453 after this change. Designs 10-12 refined the release window: design~10 added a ``near-$\gamma$'' gate to concentrate release on clauses closest to the satisfaction threshold, and design~12 restricted action to $c_m < \gamma$ (below-$\gamma$ windowed release), pushing $N_{\max}$ to 905. HEBO tuning of design~10 alone cut median steps at $N = 453$ from 45{,}234 to 12{,}409---a $3.6\times$ reduction in the recorded step count.

\subsubsection*{Phase~2: Window shaping and amplitude refinement (designs 16-285)}

With the basic release structure working up to $N \leq 905$, the search spent designs 16--285 along the primary path exploring progressively more elaborate amplitude shaping functions. Design~16 added a floor to the amplitude window. Design~20 brought in power-law shaping ($t^p$). Design~30 introduced a dead-zone threshold. Design~34 swapped the amplitude for a smoothstep function ($6t^5 - 15t^4 + 10t^3$). Design~36 applied a rational warp ($s / (s + (1-s)^2)$). These refinements pushed $N_{\max}$ to 1{,}280 (design~30), though step counts at that size remained high ($130{,}867$--$230{,}042$).

Two other directions were explored in parallel: $x_l$-dependent amplitude boosting (design~43, replacing $\mathrm{gate}_{x_l}$ with $\mathrm{gate}_{x_l} (1 + g \cdot \mathrm{gate}_{x_l})$) and $x_l$-adaptive sharpness control (designs 222--285, making the window sharpness depend on clause memory state). These additions pushed the parameter count upward (from 13 to 24) without matching gains in scaling. Design~285, the most complex solver in this phase, had 24~hyperparameters yet could only solve up to $N = 320$.

\subsubsection*{Phase~3: Simplification and component assembly (designs 298-322)}

Design~298 marks a turning point. The LLM discarded the entire accumulated complexity of Phase~2 (smoothstep, rational warp, dead zones, $x_l$-adaptive sharpness, threshold shifts) and reverted to the clean three-gate structure of design~7: $\mathcal{R}_m = \kappa \cdot \mathrm{gate}_{x_l} \cdot \mathrm{weak\_band} \cdot \mathrm{push\_up}$. The parameter count fell from 24 to 13. Reach dropped ($N_{\max}$ fell from 1{,}280 to 640), but the simplified base gave the remaining innovations a clean foundation.

From there, the system assembled the final release term's remaining components in quick succession:
\begin{itemize}
    \item Design~302: Added $\mathrm{gate}_{\mathrm{tail}}$, a sigmoid suppressing release at high $x_{s,m}$ to prevent saturation.
    \item Design~305: Added $\mu_{\mathrm{tail}} \cdot \mathrm{gate}_{x_l}$ to the tail gate, letting the safety threshold shift with $x_l$.
    \item Design~310: Added $\mathrm{amp\_norm} = x_{l,\mathrm{norm}} / (x_{l,\mathrm{norm}} + x_{l,m})$, a monotone decay preventing release amplitude from growing with $x_l$.
    \item Design~316: Flipped the tail-gate shift sign from $-\mu$ to $+\mu$, extending (not reducing) the safe range for high-$x_l$ clauses.
    \item Design~320: Squared $\mathrm{amp\_norm}$ for stronger damping at large $x_l$.
    \item Design~322: Generalized the square to a tunable power $p$, replacing $\mathrm{amp\_norm}^2$ with $\mathrm{amp\_base}^p$ and making $p$ a hyperparameter.
\end{itemize}
By design~321 (HEBO tuning of design~320), $N_{\max}$ had recovered to 1{,}280 at 118{,}834 median steps. The release term now contained all five gating components of the final solver.

\subsubsection*{Phase~4: Floor refinement and final tuning (designs 328-340)}

One problem remained: $\mathrm{amp\_norm}$ decays to zero as $x_{l,m} \to \infty$, which can starve clauses that genuinely need sustained help. Design~328 introduced an amplitude floor, $\mathrm{amp\_norm} = (1 - f) \cdot \mathrm{amp\_pow} + f$, with $f = a_{\mathrm{floor}} \cdot \mathrm{gate}_{x_l}^2$. But because the release term already multiplies by $\mathrm{gate}_{x_l}$, this created a double-gating pathology: the effective floor scaled as $\mathrm{gate}_{x_l}^3$.

Three floor variants followed in rapid succession:
\begin{itemize}
    \item Design~332: Reduced to $f = a_{\mathrm{floor}} \cdot \mathrm{gate}_{x_l}$ (linear; effective $\propto \mathrm{gate}_{x_l}^2$).
    \item Design~336: Dropped $x_l$ dependence entirely: $f = a_{\mathrm{floor}}$ (constant floor; effective $\propto \mathrm{gate}_{x_l}$).
    \item Design~339: Conditioned the floor on clause state instead: $f = a_{\mathrm{floor}} \cdot [1 - (1 - \mathrm{weak\_band})(1 - \mathrm{push\_up})]$. This soft-OR activates the floor when the clause sits in the weak-satisfaction band \emph{or} when $x_{s,m}$ is below target, and shuts off otherwise.
\end{itemize}

Design~339 reached $N_{\max} = 1{,}810$ (691{,}058 steps), the first solver in the lineage to clear the largest benchmark instances. Design~340 (HEBO tuning of 339, no code changes) then brought the median steps at $N = 1{,}810$ down from 691{,}058 to 95{,}503, a $7.2\times$ reduction in the recorded step count, illustrating the contribution of parameter tuning within the discovered architecture.

\subsubsection*{Summary: Building up the release term}

Table~\ref{tab:component_origins} records which design first introduced each component of the final release term. The core three-gate structure ($\mathrm{gate}_{x_l} \cdot \mathrm{weak\_band} \cdot \mathrm{push\_up}$) was in place by design~7; assembling the full five-gate structure with a clause-state-driven floor took 32~designs and a critical simplification event at design~298.

\begin{table}[htbp]
    \scriptsize
    \centering
    \caption{\textbf{Release term component origins.} Each row shows the first design in the lineage that introduced a component of the final release term (Eq.~\eqref{eq:release}).}
    \label{tab:component_origins}
    \begin{tabular}{llrl}
        \toprule
        Component & Symbol & Design & Short name \\
        \midrule
        Coupling strength & $\kappa$ & 5 & $x_l$-gated $x_s$ bump \\
        Long-term memory gate & $\mathrm{gate}_{x_l}$ & 5 & $x_l$-gated $x_s$ bump \\
        Weak-satisfaction band & weak\_band & 7 & Bounded $x_s$ release \\
        Bounded upward push & push\_up & 7 & Bounded $x_s$ release \\
        Tail-safety gate & $\mathrm{gate}_{\mathrm{tail}}$ & 302 & $x_s$-cap tail gate \\
        $x_l$-shifted tail & $\mu_{\mathrm{tail}}$ & 305 & $x_l$-shifted tail gate \\
        Amplitude norm & amp\_norm & 310 & Tail-safe release \\
        Tunable power & $p$ & 322 & Tunable amp$\_p$ \\
        Amplitude floor & $a_{\mathrm{floor}}$ & 328 & $x_l$-gated amp floor \\
        Clause-state floor & floor\_gate & 339 & State-gated floor \\
        \bottomrule
    \end{tabular}
\end{table}

%---------------------------------------
\subsection{Release-component ablation}
\label{sec:sm_analysis}

The release term couples clause history, current satisfaction, and short-term memory. Its three central factors favor large $x_l$, clause costs near the weak-satisfaction band, and $x_s<x_s^*$. The tail and amplitude factors further shape this intervention. The optimized values describe a selective correction: the memory gate has half-activation near $x_l=1000$, the inverted band $\eta>\gamma$ has peak amplitude about $0.16$, and the bounded push vanishes above $x_s=0.091$. These properties follow directly from the equations; they do not require an assumed distribution of visited states.

We tested the components by setting one multiplicative factor to one while keeping the optimized parameters and any internal uses of that factor unchanged. The recorded evaluation uses planted instances at $\alpha_r=4.3$, 100 instances per size, and the paper's step-budget ladder. Table~\ref{tab:gate_ablation} reports the largest size clearing the majority-solved threshold.

\begin{table}[htbp]
\centering
\caption{\textbf{Release-component ablation.} Removing the long-term-memory, weak-band, or bounded-push factor reduces reach. Removing the tail or amplitude factor preserves reach on this evaluation ladder.}
\label{tab:gate_ablation}
\begin{tabular}{lr}
\toprule
Removed multiplicative factor & $N_{\max}$ \\
\midrule
None (full design 340) & 1810 \\
$\mathrm{gate}_{x_l}$ & 905 \\
$\mathrm{weak\_band}$ & 453 \\
$\mathrm{push\_up}$ & 453 \\
$\mathrm{gate}_{\mathrm{tail}}$ & 1810 \\
$\mathrm{amp\_norm}$ & 1810 \\
\bottomrule
\end{tabular}
\end{table}

This experiment identifies the core three-factor structure as the main contributor to reach at the optimized settings. The tail and amplitude factors offer potential simplifications; their removal increases the recorded step count at $N=1810$ from 109,448 to 130,520 and 126,402, respectively, without changing the largest cleared size. The fresh comparison with a tuned baseline appears in Fig.~\ref{fig:benchmark} and Sec.~\ref{sec:benchmark}.

%%%%%%%%%%%%%%%%%%%%%%%%%%%%%%%%%%%%%%%%%%%%%%%%%%%%%%%%%%%%%%%%%%%%%%%%%%%%%%%
\section{Objective portfolio and search history}
\label{sec:sm_objectives}

The released portfolio contains 42 objectives, including the human-written initial objective, assembled from several source workspaces and subsequent work in the merged workspace. The groups below describe their formulations. 

\subsection{Objective formulations}
\label{sec:sm_phases}

\textbf{Power-law extrapolation (objectives 0--5).}
These objectives fit power-law and exponential models to median integration steps and extrapolate to $N=2000$. The initial objective uses an $R^2$-weighted blend; generated variants introduce coverage, reliability, and model-selection terms. This supplies a useful initial ranking, but small-$N$ fits can favor designs whose performance deteriorates at larger sizes.

\textbf{Censoring and restart costs (objectives 6--15).}
A second family augments the observed step count $s$ with a restart-cost surrogate,
\begin{equation}
 s_{\mathrm{eff}}=s+\frac{u}{1-u}B,
\end{equation}
where $u$ is the recorded unsolved fraction and $B$ the step budget. Other variants combine robust regression with model averaging. These formulations expose an important interface issue: successful batches terminate once $u<0.5$, so their final $u$ is a stopping statistic, not a fixed-budget estimate of the probability of failure.

\textbf{Reach-dominant metrics (objectives 16--31).}
These objectives increasingly prioritize the largest size actually cleared, with tail-scaling, headroom, and frontier reliability used to distinguish designs at similar reach. Objective 20, for example, adds milestone penalties before $N=640$ and $N=1280$ when the corresponding fidelities are available.

\textbf{Schedule-based headroom (objectives 32--41).}
The objectives generated in the merged workspace combine reach with cost relative to the prescribed schedule budget $B(N)$. They use smooth maxima or worst-window estimates to emphasize bottlenecks near the search frontier. Several also assess abrupt increases in cost across adjacent sizes or mixed outcomes in repeated measurements.

\subsection{Diagnosing misleading proxies}
\label{sec:sm_reward_hacking}

Correlation analysis gives the meta-agent a concise view of disagreement, which it combines with the recorded experiments and research goal. For example, objective 17 imposed a stricter solve-rate threshold than the evaluation's majority-solved rule and could favor broad small-size coverage over larger-size reach. Its ranking was negatively correlated with the consensus, and the recorded meta-agent assessments assigned it zero weight. Objective 32 supplies a different example: its negative median pairwise correlation causes the agreement clamp itself to set its raw weight to zero.

The meta-agent also reduced the multipliers of the early extrapolation family as the search shifted toward larger-size reach. Zero-multiplier objectives do not vote in the final weighted average, although they remain part of the portfolio used to compute pairwise agreement. Recorded multipliers must be distinguished from normalized consensus weights and from the weights active at a particular design-creation event.

\subsection{Agreement and shared bias}
\label{sec:sm_cluster_dominance}
\label{sec:sm_why_consensus}

Agreement-weighted voting suppresses isolated disagreement, but a correlated group can share the same modeling error. Meta-agent review supplies a second signal by assessing the objective definitions against the research goal. The evidence in Sec.~\ref{sec:implementation} shows both the benefit and the limit of this mechanism: later reach-based criteria receive more influence, while a large correlated block remains dominant. The repeated-run comparisons in Sec.~\ref{sec:repeated_runs} evaluate the resulting search behavior directly.

\subsection{Complete objective list}
\label{sec:sm_complete_list}
%---------------------------------------

Table~\ref{tab:sm_objectives} lists all 42~objectives with their source and final recorded meta-agent multipliers. These final multipliers were recorded after the last design and did not guide its creation; the effects of the operative multipliers are discussed in Sec.~\ref{sec:implementation}.

\begingroup
\small
\setlength{\tabcolsep}{3pt}
\begin{longtable}{rllp{10cm}r}
    \caption{\textbf{Complete list of all 42 objectives.} Each objective is listed with its ID, source workspace, formulation group, abbreviated description, and final recorded meta-agent multiplier. A zero removes its direct vote in that reweighted ranking; a dash indicates no multiplier was recorded.}
    \label{tab:sm_objectives}\\
    \toprule
    ID & Source & Group & Description & Mult. \\
    \midrule
    \endfirsthead
    \multicolumn{5}{l}{\textit{Table \thetable\ continued}} \\
    \toprule
    ID & Source & Group & Description & Mult. \\
    \midrule
    \endhead
    \multicolumn{5}{l}{\textit{Phase~1a: Power-law extrapolation}} \\
    0 & merged & 1a & $R^2$-weighted power-law / exponential blend, extrapolate to $N{=}2000$ & 0.0 \\
    1 & ws-22 & 1a & Predicted $\log_{10}$ steps at $N{=}2000$ incl.\ scaling + reliability & 0.0 \\
    2 & ws-22 & 1a & $\log_{10}$ expected steps with reliability-adjusted cost + coverage penalty & 0.0 \\
    3 & ws-22 & 1a & AIC-mixed scaling to $N{=}2000$ with smooth fail/coverage & 0.0 \\
    4 & ws-22 & 1a & Smooth reliability and budget-cap penalties & 0.0 \\
    5 & ws-22 & 1a & Scaling + reliability-margin penalties & 0.0 \\
    \midrule
    \multicolumn{5}{l}{\textit{Phase~1b: Censored-aware scaling}} \\
    6 & ws-24 & 1b & Success-rate + scaling extrapolation to $N{=}2000$ & 0.0 \\
    7 & ws-24 & 1b & Censor-aware AIC model avg.\ with failure lower-bounds & 0.0 \\
    8 & ws-24 & 1b & Success + scaling + coverage extrapolation & 0.0 \\
    9 & ws-24 & 1b & AIC-avg $\log_{10}$ expected solve-steps incl.\ success + coverage & 0.0 \\
    10 & ws-24 & 1b & AIC-avg robust scaling + reliability & 0.0 \\
    11 & ws-24 & 1b & AIC-avg $\log_{10}$ restart-steps + reliability margin & 0.0 \\
    12 & ws-24 & 1b & AIC-avg restart-steps + slope and reliability & 0.0 \\
    13 & ws-24 & 1b & $p$-adjusted steps with smooth coverage and slope penalties & 0.0 \\
    14 & ws-24 & 1b & Restart-steps with reliability margin and censoring & 0.0 \\
    15 & ws-24 & 1b & AIC-weighted censored scaling fit of restart-steps & 0.0 \\
    \midrule
    \multicolumn{5}{l}{\textit{Phase~2: Reach-dominant}} \\
    16 & ws-32 & 2 & Extrapolated log-steps + near-threshold failure penalty + reach reward & 0.0 \\
    17 & ws-33 & 2 & Composite $N_{\mathrm{eff}}$ at $u \leq 0.40$ + TTS surrogate + scaling exponent & 0.0 \\
    18 & ws-34 & 2 & Robust high-$N$ solves ($u \leq 0.3/0.2$) + brittleness penalty & 0.0 \\
    19 & ws-34 & 2 & Soft-pass area + top-level reward + anti-truncation & 0.0 \\
    20 & ws-36 & 2 & Reach-dominant: hard fail before 640/1280 + tail slope + slack & 0.2 \\
    21 & ws-36 & 2 & Reach + budget-margin/slack + conservative next-$N$ clearance & 0.3 \\
    22 & ws-36 & 2 & Successful-prefix clearance $N^*$ + near-budget tail penalty & 0.4 \\
    23 & ws-36 & 2 & Reach + conservative worst-of-multi-fit clearance at 1280/2560 & 0.0 \\
    24 & ws-36 & 2 & Reach + worst-window predicted earliest schedule failure & 0.3 \\
    25 & ws-36 & 2 & Robust scheduled $N^*$ with ratio-space clearance $\leq 0.60$ & 0.0 \\
    26 & ws-36 & 2 & Gated tail-only headroom-to-1280 + slope/curvature penalties & 0.0 \\
    27 & ws-36 & 2 & Tail-local polynomial scaling + budget-clearance robustness & 0.0 \\
    28 & ws-36 & 2 & Two-jump headroom at 640$\to$905$\to$1280 + exponent penalty & 0.0 \\
    29 & ws-37 & 2 & Reach + low scaling exponent + censored extrapolation to $N{=}2000$ & 0.0 \\
    30 & ws-37 & 2 & Reach + scaling + headroom + frontier robustness & 0.0 \\
    31 & ws-37 & 2 & Schedule-faithful lexicographic: reach, then steps, then slope & 0.1 \\
    \midrule
    \multicolumn{5}{l}{\textit{Phase~3: Schedule-faithful}} \\
    32 & merged & 3 & Tail headroom: conservative worst-window next-$N$ overshoot & 0.0 \\
    33 & merged & 3 & Lexicographic reach + strong tail headroom tie-break & 1.2 \\
    34 & merged & 3 & Reach + tail headroom + worst-window 1810/2560 overshoot & 2.0 \\
    35 & merged & 3 & Reach + schedule-budget headroom + budget-cliff hazard & 2.2 \\
    36 & merged & 3 & Reach + smooth-max headroom + hazard + repeat-stability & 1.2 \\
    37 & merged & 3 & Reach + smooth-max headroom + focus 1280$\to$1810$\to$2560 + hazard & 2.4 \\
    38 & merged & 3 & Reach + smooth-max headroom + tail exponent + post-1280 cliff & 1.6 \\
    39 & merged & 3 & Reach + smooth-max $\log$(headroom) + multi-jump hazard + censor & 3.0 \\
    40 & merged & 3 & Reach + smooth-max $\log$(headroom) + cliff hazard + robustness & 3.2 \\
    41 & merged & 3 & Three-variant objective (balanced / hazard / robustness) & --- \\
    \bottomrule
\end{longtable}
\endgroup

%%%%%%%%%%%%%%%%%%%%%%%%%%%%%%%%%%%%%%%%%%%%%%%%%%%%%%%%%%%%%%%%%%%%%%%%%%%%%%%
\section{Workflow development and workspace merging}
\label{sec:sm_methodology}
\label{sec:sm_meta_agent}
\label{sec:sm_schedule}
\label{sec:sm_workspace_merging}

The framework separates objective generation from strategic review: the objective agent proposes scoring functions, and the meta-agent reviews their behavior and assigns multipliers. The evaluation schedule remains fixed during a run, with rule-based promotion to higher fidelities. This gives each design a common ladder of problem sizes and budgets while concentrating evaluation on promising candidates (Sec.~\ref{sec:protocol}).

The exploratory campaign began in separate workspaces. Selected designs were imported with their parent-child references, preserving the context used by the planner. The merged graph lets the planner build on candidates from different workspaces, and the released database contains the resulting design genealogy. The imported records retain their source-workspace counters, so these counters should be interpreted locally when reconstructing objective age or planner history.

%%%%%%%%%%%%%%%%%%%%%%%%%%%%%%%%%%%%%%%%%%%%%%%%%%%%%%%%%%%%%%%%%%%%%%%%%%%%%%%
\section{Base Solver Framework}
\label{sec:sm_solver_framework}
%%%%%%%%%%%%%%%%%%%%%%%%%%%%%%%%%%%%%%%%%%%%%%%%%%%%%%%%%%%%%%%%%%%%%%%%%%%%%%%

This section describes the baseline DMM solver provided to the LLM agents and the modular code framework that supports automated solver generation.

%---------------------------------------
\subsection{Research goal and problem context}
\label{sec:sm_research_goal}
%---------------------------------------

The high-level research goal given to the meta-agent is:
\begin{quote}
\emph{Develop an algorithm that efficiently solves hard, large-scale 3-SAT problems. Prioritize robust scaling, with focus on large-$N$ behaviors. Aim for polynomial step vs.\ $N$ scaling, ideally sub-quadratic. Common pitfall: polynomial at small $N$, exponentially diverging at large $N$.}
\end{quote}

\paragraph{3-SAT and the DMM relaxation.}
A 3-SAT instance has $N$ Boolean variables $V_i \in \{0,1\}$ and $M$ clauses $C_m = L_i \vee L_j \vee L_k$, where each literal $L_i$ is either $V_i$ or its negation. A digital MemComputing machine (DMM)~\citep{diventra_memcomputing_2022,bearden_efficient_2020} relaxes each $V_i$ to a continuous variable $v_i \in [-1, 1]$ and introduces auxiliary \emph{memory} variables that encode information about the dynamics' history. The sign of $v_i$ supplies a candidate Boolean assignment, checked against every clause during integration.

\paragraph{Baseline DMM equations.}
The baseline dynamics (Eqs.~\eqref{eq:dmm_v}--\eqref{eq:dmm_xl} of the main text) evolve three sets of variables. The state variables $v_n$ follow
\begin{equation}
    \dot{v}_n = \sum_{m=1}^{M} \left[x_{l,m}\, x_{s,m}\, G_{n,m} + (1 + \zeta\, x_{l,m})(1 - x_{s,m})\, R_{n,m}\right],
    \label{eq:sm_baseline_v}
\end{equation}
where $G_{n,m}$ is a gradient term nudging variables toward clause satisfaction and $R_{n,m}$ is a rigidity term preventing satisfied literals from flipping:
\begin{align}
    G_{n,m} &= \tfrac{1}{2}\, q_{n,m} \min\bigl[(1 - q_{j,m} v_j),\, (1 - q_{k,m} v_k)\bigr], \label{eq:sm_G} \\
    R_{n,m} &= \begin{cases}
        \tfrac{1}{2}\, q_{n,m}(1 - q_{n,m} v_n), & \text{if } c_m = \tfrac{1}{2}(1 - q_{n,m} v_n), \\
        0, & \text{otherwise},
    \end{cases} \label{eq:sm_R}
\end{align}
with $q_{n,m} \in \{+1, -1\}$ the literal polarity and the clause cost
\begin{equation}
    c_m = \tfrac{1}{2}\min\bigl[(1 - q_{i,m} v_i),\, (1 - q_{j,m} v_j),\, (1 - q_{k,m} v_k)\bigr].
    \label{eq:sm_cm}
\end{equation}
Two auxiliary memory variables per clause track satisfaction history:
\begin{align}
    \dot{x}_{s,m} &= \beta\, (x_{s,m} + \epsilon)\,(c_m - \gamma), \label{eq:sm_xs} \\
    \dot{x}_{l,m} &= \alpha\, (c_m - \delta). \label{eq:sm_xl}
\end{align}
The short-term switch $x_{s,m} \in [0,1]$ toggles between ``push'' ($x_s \approx 1$) and ``hold'' ($x_s \approx 0$) modes; the long-term weight $x_{l,m} \in [1, 10^6]$ increases when $c_m>\delta$ and decreases when $c_m<\delta$, subject to its bounds. Default hyperparameters are $\alpha = 5$, $\beta = 20$, $\gamma = 0.25$, $\delta = 0.05$, $\epsilon = 10^{-3}$, and $\zeta = 10^{-3}$.

%---------------------------------------
\subsection{Modularized solver framework}
\label{sec:sm_modular_framework}
%---------------------------------------

The accompanying solver codebase separates problem-specific dynamics from generic solving infrastructure. A single Python file fully defines each solver experiment through three components:

\begin{enumerate}
    \item \textbf{\texttt{VARIABLES\_SPEC}}: a dictionary mapping variable names to their initialization, shape, and bounds. For the baseline:
    \begin{itemize}
        \item \texttt{v}: shape $(B, N)$, bounds $[-1, 1]$, randomly initialized;
        \item \texttt{xl}: shape $(B, M)$, bounds $[1, 10^6]$, initialized to~1;
        \item \texttt{xs}: shape $(B, M)$, bounds $[0, 1]$, initialized to~0;
    \end{itemize}
    where $B$ is the batch size. The LLM can add, remove, or modify variables to introduce new memory mechanisms.

    \item \textbf{\texttt{HYPER\_SPACE}}: a dictionary defining the hyperparameter search space. Each entry specifies a type (uniform, log-uniform, integer, or categorical), default value, and bounds. The baseline defines seven parameters ($\alpha, \beta, \gamma, \delta, \epsilon, \zeta$, and the integration step size). A Bayesian optimizer (HEBO \citep{cowen-rivers_hebo_2022}) periodically tunes these parameters for selected solver designs.

    \item \textbf{\texttt{\_grad\_single(vars, idx, sgn, hp)}}: the core dynamics function computing per-instance gradients for all state variables. Inputs are the current variable values (\texttt{vars}), clause structure (\texttt{idx}, \texttt{sgn}), and hyperparameters (\texttt{hp}). It returns a gradient dictionary with the same keys as \texttt{vars}. This function encodes the dynamical equations (Eqs.~\eqref{eq:sm_baseline_v}-\eqref{eq:sm_xl}) and is the primary target of LLM-driven design.
\end{enumerate}

\paragraph{Framework integration.}
The solver framework (\texttt{sat\_solver.py}) dynamically imports these three components at runtime. Launching an experiment with solver ID $k$ loads \texttt{solvers/solver\_$k$.py}. The \texttt{SATSolver} class reads \texttt{VARIABLES\_SPEC} to allocate and initialize state tensors, then vectorizes \texttt{\_grad\_single} over the batch dimension via \texttt{torch.vmap} \citep{paszke_pytorch_2019}. The solving loop is standard: zero gradients, compute dynamics through the vmapped function, take an optimizer step, clamp variables to their specified bounds, check for satisfying assignments.

\paragraph{LLM-generated solvers.}
The designer agent produces new solver files containing modified versions of these three components. Because the interface is fixed---any file exporting \texttt{VARIABLES\_SPEC}, \texttt{HYPER\_SPACE}, and \texttt{\_grad\_single} with the correct signatures is automatically integrated---the LLM can freely redesign the dynamics, introduce new auxiliary variables, or restructure the hyperparameter space without touching framework code. This modularity is what makes the automated search described in the main text possible: each MCGS iteration generates a new solver file, evaluates it through the framework, and records the results.

%%%%%%%%%%%%%%%%%%%%%%%%%%%%%%%%%%%%%%%%%%%%%%%%%%%%%%%%%%%%%%%%%%%%%%%%%%%%%%%
\section{LLM Agent Prompts}
\label{sec:sm_prompts}

This section preserves the agent prompt templates accompanying the original campaign, organized by role. The executable templates and configuration are included in the code archive.

%------------------------------------------------------------------------------
\subsection{Planner Agent}
\label{sec:sm_prompts_planner}

The Planner Agent analyzes research progress and assigns strategic directions to Designer Agents. It operates in two stages: first selecting promising experiments to review from MCGS rankings, then synthesizing insights into distinct, non-overlapping research directions.

\paragraph{System prompt.}
Sets the agent's role and scope.

\begin{lstlisting}[style=prompt]
You are the Planner Agent in an LLM-driven autonomous
research system. Guide the research direction and provide
strategic recommendations for Designer Agents.
\end{lstlisting}

\paragraph{Stage 1: Direction selection.}
The planner receives an overview of all experiments ranked by UCB score from MCGS and selects which experiments to examine in detail.
\begin{lstlisting}[style=prompt]
## Research Context
{main_research_context}

## Task
Select {DESIGNER_AGENT_COUNT} promising research
directions and identify experiments to review in detail.

## Database Overview
- Total experiments: {total_experiments}
- Best performance: {best_performance}
- Recent experiments: {recent_summary}

## Experiment Summaries
Up to {MAX_EXPERIMENT_SUMMARY} experiments ranked by
upper confidence bound (UCB) from Monte Carlo graph
search (MCGS):
{experiment_summaries}

## Guidelines
- Request details for up to {MAX_EXPERIMENT_DETAIL}
  experiments
- Prioritize high-UCB experiments and complementary ideas
- Avoid redundant or near-duplicate directions
- Hyperparameter tuning is automatic via HEBO optimizer -
  don't assign this to Designer Agents
- Your goal: MINIMIZE the objective function

## Output Format
```json
{
  "lookup_experiment_ids": [/* int IDs to review */],
  "context_rationale": "Concise reasoning behind chosen
    directions and their relevance"
}
```
\end{lstlisting}

\paragraph{Stage 2: Designer assignment.}
After reviewing the requested experiment details, the planner synthesizes findings into non-overlapping research directions with reference experiments for each Designer Agent.
\begin{lstlisting}[style=prompt]
## Experiment Details
{design_details}

## Task
Summarize progress and assign {DESIGNER_AGENT_COUNT}
designer agents distinct research directions with up to
{MAX_DESIGNER_REFERENCE} reference experiments each.

## Output Format
```json
{
  "current_phase": "early_exploration|systematic_search
    |exploitation|stagnation|breakthrough_needed",
  "key_insights": ["Top takeaways explaining what works"],
  "success_patterns": [
    "Shared traits of strong experiments"],
  "failure_patterns": [
    "Shared traits of weak experiments"],
  "research_directions": [
    "Direction for d1", "Direction for d2"],
  "strategy_rationales": [
    "Rationale for d1", "Rationale for d2"],
  "focus_areas": ["themes for d1", "themes for d2"],
  "avoid_areas": ["pitfalls for d1", "pitfalls for d2"],
  "reference_design_ids": [
    [int IDs for d1], [int IDs for d2]]
}
```

## Guidelines
- All arrays must have length {DESIGNER_AGENT_COUNT}
  and align by index
- Directions must be non-overlapping, complementary,
  and distinct
\end{lstlisting}

%------------------------------------------------------------------------------
\subsection{Designer Agent}
\label{sec:sm_prompts_designer}

The Designer Agent creates new experiment designs based on the Planner's strategic directions and reference experiments. It operates through a multi-turn conversation: first proposing a design, then receiving experiment results, and finally analyzing outcomes. If execution errors occur, an error-recovery prompt is used.

\paragraph{System prompt.}
Sets the agent's role and scope.

\begin{lstlisting}[style=prompt]
You are the Designer Agent in an LLM-driven autonomous
research system. You receive strategic recommendations
from the Planner Agent and create targeted experiments.
\end{lstlisting}

\paragraph{Design creation.}
The designer receives the research context, planner guidance, baseline components, and reference experiments, then proposes a new design with one principled modification.
\begin{lstlisting}[style=prompt]
## Research Context
{main_research_context}

## Planner Context
{planner_context}

## Architecture
- `domain_knowledge/{framework_module}.py` - Main
  framework
- `domain_knowledge/{baseline_filename}` - Baseline
  components
- `solvers/solver_N.py` - Your experiment components
- Objective: Consensus of all generated objectives
  (to MINIMIZE)
- `schedules/schedule_{current_schedule_id}.py` - Current
  experiment schedule

{framework_module}.py dynamically imports your
{num_components} components from solver_N.py.

## Baseline Components
```python
{base_solver_code}
```

## Reference Experiments
{reference_experiments}

## Current Objective
{objective_description}

## Current Schedule
{schedule_description}

## Task
Design a new experiment by modifying the {num_components}
core components:
1. Make ONE small, principled modification to baseline
2. Build on proven ideas from reference experiments
3. Follow the Planner's direction and rationale
4. Explain how changes should improve the objective

## Required Components
{component_descriptions}

Available imports: `math`, `numpy`, `scipy`, `torch`,
and standard libraries

## Output Format
Return a JSON object with:
```json
{
  "explanation": "Rationale for modification, referencing
    evidence and strategy",
  "solver_code": "Complete Python code with imports and
    components: {component_names}"
}
```
\end{lstlisting}

\paragraph{Experiment results.}
After the design is executed, the experiment results are appended to the conversation as context. No LLM response is requested at this stage.
\begin{lstlisting}[style=prompt]
## Experiment Results

You have completed experiments at **{current_fidelity}**
fidelity.

### Results
```json
{experiment_results}
```

**Objective value** (lower is better): {objective_value}
\end{lstlisting}

\paragraph{Result analysis.}
The designer analyzes outcomes and extracts actionable insights. Reference weights are computed for MCGS graph updates, reflecting how much each parent design influenced the current result.
\begin{lstlisting}[style=prompt]
## Task
Analyze results and extract actionable insights. Focus on
why the outcome occurred and what to do next.
Evaluate how much each reference design influenced this
result for Monte Carlo Graph Search (MCGS) updates.

## Success Levels
- **excellent**: Major breakthrough or validated
  improvement
- **good**: Noticeable improvement with well-understood
  cause
- **moderate**: Partial progress or useful insight
  despite limited gains
- **poor**: No improvement or regression, but still
  informative

## Reference Weights
- Include only referenced design IDs
- Each weight (0-1) represents influence on current
  design
- Weights must sum to 1.0
- Higher weights for ideas/parameters that most strongly
  shaped results

## Output Format
Return a JSON object with:
```json
{
  "short_name": "Concise descriptive title (<= 40 chars)",
  "key_insight": "Most important takeaway (1 line)",
  "success_level": "poor|moderate|good|excellent",
  "detailed_analysis": "Comprehensive explanation of
    mechanisms and outcomes",
  "comparison_to_references": "How results compare with
    referenced designs",
  "recommended_next_steps": "Concrete suggestions for
    future designs",
  "reference_weights": [
    {"design_id": int, "weight": float}
  ]
}
```
\end{lstlisting}

\paragraph{Error recovery.}
When a design produces runtime errors, this prompt is appended to the existing conversation so the designer can see the full context of the failed attempt.
\begin{lstlisting}[style=prompt]
## Error
```
{error_message}
{full_traceback}
```

## Task
Fix the error in your implementation components and
return all {num_components} corrected components:
{component_names}.

## Output Format
Return a JSON object with:
```json
{
  "error_summary": "What went wrong and how you fixed it",
  "solver_code": "Complete Python code with imports and
    corrected components"
}
```
\end{lstlisting}

%------------------------------------------------------------------------------
\subsection{Objective Agent}
\label{sec:sm_prompts_objective}

The Objective Agent generates proxy objective functions that guide the search. The evaluation schedule remains fixed at the baseline; only the objective function is generated. The Meta-Agent provides strategic guidance on what properties the next objective should emphasize.

\paragraph{System prompt.}
Sets the agent's role and scope.

\begin{lstlisting}[style=prompt]
## Role
You are the **Objective Agent**. You design proxy
objective functions that guide the discovery of better
algorithms.
\end{lstlisting}

\paragraph{Objective generation.}
The Objective Agent receives the current research state, Meta-Agent guidance, baseline code, existing objectives, and recent results, then proposes a new proxy objective function.
\begin{lstlisting}[style=prompt]
## Research Context
{main_research_context}

## Meta-Agent Guidance
{meta_agent_directions}

Consider this guidance when designing your objective
function. The meta-agent has analyzed research progress
and identified areas that need attention.

## Task
Generate a proxy objective function that better estimates
the research goal. The experiment schedule will use the
baseline schedule (shown below for reference).

## Discovery Philosophy
The baseline objective is deliberately simple. You have
freedom to design objectives that:
- Target any experiment or combination of all experiments
- Use any combination of available metrics
- Apply any scaling model or none at all
- Incorporate uncertainty, robustness, or other advanced
  concepts

Learn from existing results and think about what truly
matters for identifying the best algorithms.

## Baseline Code
**Objective:**
```python
{baseline_objective_code}
```

**Schedules (for reference - will be used unchanged):**
```python
{baseline_schedule_code}
```

## Existing Objectives
{existing_objective_summary}

## Recent Experiment Results
{recent_experiments_objectives_summary}

## Required Function

**Objective function:**
```python
def objective(experiment_results):
    """Estimate the research goal from experiment results.

    Args:
        experiment_results: List of dicts with experiment
            details
    Returns:
        Float to be MINIMIZED
    """
    return objective_value
```

## Experiment Interface
```python
{experiment_code}
```

## Guidelines
- `experiment_results` is a list of dicts containing all
  experiment kwargs and outputs
- Objective should be smooth and friendly to Bayesian
  optimization (avoid large penalties for failures)
- Objective should adapt to different schedules and remain
  backward compatible when possible
- Available imports: `math`, `numpy`, `scipy`, `torch`,
  and standard libraries

## Output Format

Return a JSON object with:
```json
{
  "objective_description": "What this objective measures
    (one line, comprehensive but extremely concise)",
  "objective_code": "Complete Python code with imports and
    objective() function"
}
```
\end{lstlisting}

\paragraph{Error recovery.}
Appended to the conversation when the generated objective function produces a runtime error.
\begin{lstlisting}[style=prompt]
## Error
```
{error_message}
{full_traceback}
```

## Task
Fix the error in your objective code.

## Output Format

Return a JSON object with:
```json
{
  "error_summary": "What went wrong and how you fixed it",
  "objective_code": "Corrected Python code with imports
   and objective() function"
}
```
\end{lstlisting}

%------------------------------------------------------------------------------
\subsection{Meta-Agent}
\label{sec:sm_prompts_meta}

The Meta-Agent oversees the entire research process. It analyzes objective function performance using Kendall tau correlations, adjusts objective weights in the consensus mechanism, and provides strategic guidance for the Objective Agent's next generation.

\paragraph{System prompt.}
Sets the agent's role and scope.

\begin{lstlisting}[style=prompt]
You are the **Meta-Agent**. You oversee the entire
research process, analyze what's working and what's not,
guide the objective agent, and adjust objective weights
to improve research progress.
\end{lstlisting}

\paragraph{Research analysis.}
The Meta-Agent receives a comprehensive view of all objective functions, their correlation structure, weighting state, and recent progress, then produces an assessment with updated weights and directions for the Objective Agent.
\begin{lstlisting}[style=prompt]
## Research Context
{main_research_context}

## High-Level Research Goal
{high_level_research_goal}

## Task
Analyze research progress, evaluate objective functions,
and provide strategic guidance.

Your responsibilities:
1. Assess research progress
2. **Maintain and update an evolving consensus objective
   function**
   - Objective functions are periodically generated by
     Objective Agent
   - Planner/Designer Agents/hyperparameter optimizer
     minimize the consensus objective.
   - You can adjust objective weights: amplify useful
     ones, suppress harmful ones.
3. Guide the Objective Agent in generating new objectives.

## Experiment Schedule
(for reference - will be used unchanged):
```python
{baseline_schedule_code}
```

## Current Objective Functions

{objective_summary_with_code}

## Objective Performance Analysis

**Kendall Tau Correlation Matrix** (measures agreement
between objectives):
{objective_correlation_matrix}

## Objective Weighting Mechanism

```python
weight = default_weight * weight_multiplier
# You assign weight_multiplier, default 1.0
default_weight = agreement * age_decay
agreement = max(median_tau, 0.0)
age_decay = 0.9 ** rounds_since_creation
```

**Objective Agreement Summary**:
{objective_agreement_summary}

**Recent Progress**:
{recent_progress_summary}

**Top Designs**:
{top_designs_summary}

## Previous Meta-Agent Guidance

{previous_meta_guidance}

## Guidelines

**For Objective Evaluation**:
- Objectives with low Kendall tau (disagreeing with others)
  may be misleading
- Objectives that rank failed designs highly are
  problematic
- Objectives that don't differentiate between designs are
  not useful
- Consider whether the objective aligns with the
  high-level research goal

## Output Format

Return a JSON object:
```json
{
  "research_assessment": "Overall assessment of research
    progress (2-4 sentences)",
  "research_phase": "exploring|converging|stuck
    |breakthrough_needed|refining",
  "objective_analysis": [
    {
      "objective_id": 0,
      "assessment": "How this objective is performing",
      "weight_multiplier": 1.0
    }
  ],
  "objective_directions": "Strategic guidance for next
    objective generation (be specific)"
}
```
\end{lstlisting}

\end{document}